\documentclass[acmtog]{acmart}
\AtBeginDocument{%
  }

\copyrightyear{2026}
\acmYear{2026}
\setcopyright{cc}
\setcctype{by}
\acmConference[SA Conference Papers '26]{SIGGRAPH Asia 2026 Conference Papers}{December 01--04, 2026}{Kuala Lumpur, Malaysia}
\acmBooktitle{SIGGRAPH Asia 2026 Conference Papers (SA Conference Papers '26), December 01--04, 2026, Kuala Lumpur, Malaysia}
\acmDOI{10.1145/3829340.3842307}
\acmISBN{979-8-4007-2842-6/2026/12}

\usepackage{colortbl}
\usepackage{multirow}
\usepackage{enumitem}

\newcommand{\SysName}[0]{\textbf{SketchFlow}}

\definecolor{zjcl}{RGB}{128, 0, 128}

\usepackage{xspace}

\newcommand{\eg}{\textit{e.g.}\xspace}

\begin{document}

\title{\SysName: Zero-Shot Vector Sketch Generation via GMM Prior Flow in CLIP Latent Space}
\author{Jin Zhou}
\authornote{Jin Zhou and Hongliang Yang contributed equally to this work.}
\email{doudin2618@gmail.com}
\orcid{0009-0009-2948-3150}

\author{Hongliang Yang}
\authornotemark[1]
\email{hongliang.cscg@gmail.com}
\orcid{0009-0008-8530-9125}

\author{Pengfei Xu}
\authornote{Corresponding author: Pengfei Xu}
\email{xupengfei.cg@gmail.com}
\orcid{0000-0003-4770-4374}

\author{Hui Huang}
\email{hhzhiyan@gmail.com}
\orcid{0000-0003-3212-0544}

\affiliation{%
 \department{Guangdong Provincial Key Laboratory of Visual Media and Multidimensional Intelligence, CSSE}
  \institution{Shenzhen University}
  \city{Shenzhen}
  \country{China}
}

\renewcommand{\shortauthors}{Jin Zhou, Hongliang Yang, Pengfei Xu, and Hui Huang}

\begin{teaserfigure}
    \centering
    \includegraphics[width=\textwidth]{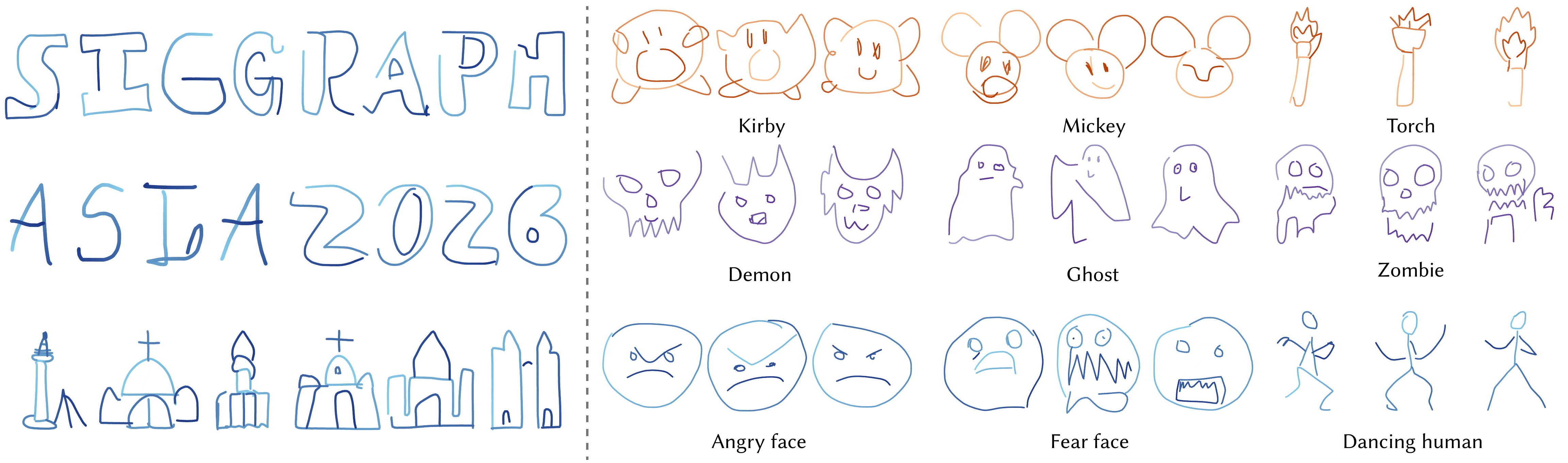}
    \Description{A grid of computer-generated sketches showing a composite scene on the left and various characters like Kirby and Mickey on the right.}
    \caption{
        \textbf{Local zero-shot vector sketch generation by \SysName{}.} \SysName{} leverages drawing priors from the CLIP latent space to synthesize prompts beyond its 345-class QuickDraw training vocabulary. Left: Text-prompted generation of letters, numbers, and landmarks in Malaysia. Right: Diverse zero-shot generation including pop-culture characters, creatures, abstract emotions, and dynamic poses. \SysName\ produces visually appealing sketches with human-like drawing behaviors without category-specific training examples for these prompts.
    }
    \label{fig:teaser}
\end{teaserfigure}

 \begin{abstract}

Vector sketches remain one of the most concise and immediate mediums for abstract human expression. However, generating high-quality vector strokes that exhibit human-like drawing styles remains an open challenge due to the severe scarcity of fine-grained, high-quality text-to-sketch paired data. Existing text-conditioned generation methods often rely on unstable, time-consuming optimization or struggle to generalize to unseen categories in a zero-shot manner. To address these limitations, we present \SysName{}, a novel generative framework rooted in Optimal Transport (OT) theory and flow matching. By leveraging pre-trained CLIP models to bypass labor-intensive image-level text annotations, we formulate cross-modal alignment as a continuous mapping problem directly within the CLIP latent space. To bridge the inevitable modality gap between discrete text concepts and continuous sketch features, we first inject noise into discrete category embeddings to construct a continuous Gaussian Mixture Model (GMM) prior. We then utilize an Optimal Transport Conditional Flow Matching (OT-CFM) model to learn a deterministic vector field mapping from this continuous GMM prior to the target sketch feature distribution. Finally, a Hybrid Diffusion Decoder, fusing 1D U-Net and Transformer architectures, is designed to decode these features into fast and high-fidelity stroke trajectories. Extensive experiments demonstrate that \SysName{} substantially outperforms existing baselines in visual quality and adherence to natural human drawing styles. Furthermore, our geometry-preserving framework demonstrates promising local zero-shot synthesis for prompts beyond the QuickDraw training vocabulary, including unseen concept labels and semantic modifiers, while enabling smooth, continuous semantic interpolation between distinct concepts. Source code is available at: https://github.com/doudin404/SketchFlow.

\end{abstract}



\maketitle

\section{Introduction}

Vector sketches are a concise, direct medium for abstract human expression; text-conditioned vector sketch synthesis therefore has significant potential for digital avatar creation, conceptual design, and interactive art. Yet generating high-quality vector strokes with human-like drawing styles remains an open challenge. Optimization-based approaches~\cite{xing2023diffsketcher,arar2025swiftsketch, polaczek2025neuralsvg} are slow, unstable, and prone to noise. Raster-based methods~\cite{hu2024scale} lack stroke-topology constraints and introduce artifacts such as inconsistent thickness, semi-transparency, and unintended branching that impede vector tracing. Direct sequence generation with Large Language Models (LLMs)~\cite{vinker2025sketchagent} bypasses raster-image generation but produces trajectories that deviate significantly from authentic human drawing habits in style and abstract logic.

The most intuitive approach to this task is end-to-end training using large-scale, paired text-vector sketch data. However, high-quality annotated data of this nature is exceedingly scarce. Although QuickDraw~\cite{ha2017neural}, currently the largest vector sketch dataset, contains a massive collection of human-drawn trajectories, it provides only 345 discrete category labels and lacks fine-grained textual descriptions.

Consequently, we confront a fundamental research gap: How can we achieve zero-shot text-to-sketch synthesis for visual concepts absent from a fixed training vocabulary using only sketches with discrete category labels? To address this, we present \SysName{}, a novel generative framework rooted in Optimal Transport theory and flow matching, which constructs continuous, deterministic trajectories between distributions. By formulating cross-modal alignment as a continuous mapping problem, \SysName{} enables zero-shot vector sketch generation via a GMM prior flow embedded directly in the CLIP latent space. Here, \emph{zero-shot} denotes generation from prompts whose target concept labels are absent from the 345-class QuickDraw training vocabulary, leveraging the semantic structure encoded by the pre-trained CLIP model.

Our core insight is to use pre-trained CLIP to bypass image-level text annotation and align text with sketches in a shared conceptual space. We formulate this task as a cross-modal latent space mapping problem. Raw text embeddings remain mismatched with rendered-sketch embeddings, so directly conditioning the decoder on them often fails (Figure~\ref{fig:exp_ablation_flow}). We therefore expand discrete category embeddings into a continuous GMM prior, providing a smooth source distribution for flow matching, and learn an OT-CFM vector field that transports this prior to the sketch-feature distribution. A hybrid diffusion decoder combining a 1D sequence-adapted U-Net and Transformer then maps the transported features to high-fidelity stroke trajectories.

Extensive experimental results demonstrate the superiority of our approach. This work provides a practical framework for vector sketch generation and highlights the potential of flow matching techniques in cross-modal zero-shot generation. Our core contributions are summarized as follows:

\begin{itemize}[leftmargin=*, nosep]
    \item \textbf{CLIP-mediated local zero-shot synthesis:} Our model generates recognizable sketches for unseen concept labels and semantic modifiers beyond the QuickDraw training vocabulary. The results demonstrate promising local generalization through the semantic structure encoded in the pre-trained CLIP latent space.
    \item \textbf{Novel cross-modal mapping framework:} We introduce an Optimal Transport Conditional Flow Matching model combined with a continuous GMM prior to effectively bridge the modality gap between discrete text concepts and continuous sketch features in the CLIP latent space.
    \item \textbf{High-fidelity stroke reconstruction:} We design a hybrid diffusion decoder fusing 1D U-Net and Transformer architectures, which outperforms existing baselines in visual quality and adherence to human drawing styles.
\end{itemize}

 \begin{figure*}[t]
    \centering
    \includegraphics[width=\textwidth]{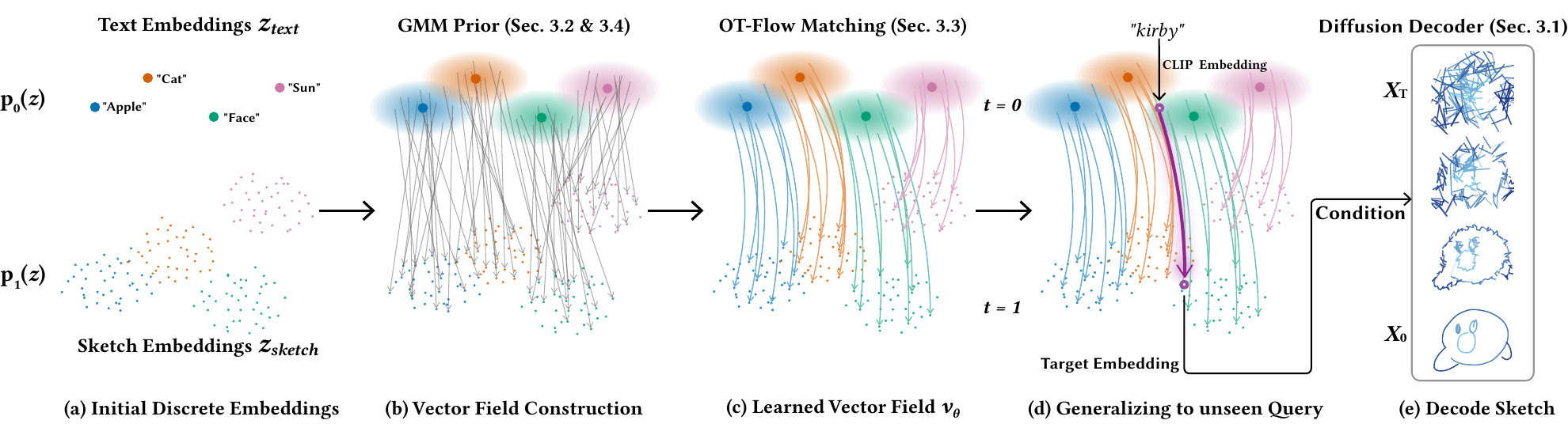}
    \caption{\textbf{Overview of the \SysName{} generative framework.} 
    (a) \textbf{Initial Discrete Embeddings:} The framework extracts discrete text embeddings $z_{\text{text}}^k$ and sketch embeddings for known categories in the latent space. 
    (b) \textbf{Vector Field Construction:} A noise injection mechanism is introduced to transform sparse concept anchors into a continuous probability density field, establishing a Gaussian Mixture Model (GMM) prior (Sec. 3.2 \& 3.4). 
    (c) \textbf{Learned Vector Field $v_\theta$:} Optimal Transport Conditional Flow Matching establishes a deterministic mapping between the GMM prior $p_0(z)$ and the target sketch distribution, naturally yielding straight-path trajectories (Sec. 3.3). 
    (d) \textbf{Generalizing to an Unseen Query:} During inference, the CLIP embedding of an unseen query (e.g., "kirby") follows the learned transport field, informed by known training concepts, toward the rendered-sketch CLIP feature distribution.
    (e) \textbf{Decode Sketch:} Conditioned on the acquired target sketch embedding, a Hybrid Diffusion Decoder utilizes a reverse diffusion process to decode pure noise $X_T$ into a final high-fidelity vector stroke sequence $X_0$ (Sec. 3.1).}
    \Description{Five-stage pipeline diagram. Discrete CLIP text anchors and rendered-sketch embeddings form colored clusters; noise expands the text anchors into a GMM prior, and OT flow matching learns non-crossing paths to the sketch distribution. At inference, a Kirby query follows a nearby transport path to a target embedding that conditions the diffusion decoder, which converts noise into a vector sketch.}
    \label{fig:pipeline_overview}
\end{figure*}

\section{Related Work}

\subsection{Vector Sketch Generation}
Vector sketches possess an inherent temporal structure; thus, early research naturally formulated the generation process as trajectory prediction over ordered strokes. SketchRNN~\cite{ha2017neural} pioneered this approach by introducing the QuickDraw dataset alongside an LSTM-VAE architecture for sketch encoding and decoding. Subsequent works~\cite{aksan2020cose,das2021sketchode,wang2023sketchknitter,das2023chirodiff,xu2026sketchformerpp,bandyopadhyay2024sketchinr,ashcroft2023strokecloud} have refined stroke modeling by utilizing RNNs, Ordinary Differential Equations (ODEs), and sequence diffusion. While these models effectively capture temporal dynamics, they typically support only unconditional or few-shot conditional generation, which severely limits their scalability and text controllability. Although Scale-Adaptive Diffusion~\cite{hu2024scale} demonstrated large-scale training capabilities on the QuickDraw dataset, it generates raster images rather than scalable vector sketches. More recently, StrokeFusion~\cite{zhou2026strokefusion} proposed training a diffusion model to denoise and generate stroke sets; however, it requires a manually specified upper bound for the number of strokes and struggles to generalize to unseen categories in a zero-shot manner.

\subsection{Text-Conditioned Visual Synthesis}
Controllable visual synthesis accepts intuitive inputs such as text, sketches, and strokes~\cite{xue2022deep,ma2024clipflow,bao2025multiway}. CLIPDraw, CLIPasso, and CLIPascene optimize B\'ezier curves through differentiable rendering to maximize text-image similarity~\cite{li2020diffvg,frans2022clipdraw,vinker2022clipasso,vinker2023clipascene}. Without ground-truth sketch references, these methods do not reproduce authentic hand-drawn style, and direct CLIP optimization is unstable.

Diffusion-prior approaches include DiffSketcher (latent diffusion and SDS), SVGDreamer and SVGDreamer++ (vector-particle optimization), SwiftSketch (stroke-order constraints), style customization, and NeuralSVG (implicit curves)~\cite{xing2023diffsketcher,poole2022dreamfusion,xing2024svgdreamer,xing2025svgdreamerpp,arar2025swiftsketch,zhang2025style,polaczek2025neuralsvg}. Diffusion priors also support sketch extraction~\cite{yun2026diffsketch}. These methods optimize predefined vector primitives, an initialization ill-suited to sparse hand-drawn sketches with few smooth, long strokes; SketchFlow instead learns stroke structure from human trajectories.

Language-model-based vector generation includes off-the-shelf models and models trained or fine-tuned on paired data. SketchAgent~\cite{vinker2025sketchagent} uses an off-the-shelf multimodal LLM: it provides broad semantics and logical drawing order but does not learn human drawing conventions, often producing rigid SVG graphics. IconShop, StarVector, OmniSVG, and LLM4SVG~\cite{wu2023iconshop,rodriguez2025starvector,yang2025omnisvg,xing2025llm4svg} are trained or fine-tuned on paired condition--SVG data, making generalization depend on training-condition coverage. SketchFlow instead learns human drawing behavior from category-labeled sketches and uses CLIP-space transport to generalize beyond the training vocabulary without diverse text descriptions.

\subsection{Flow Matching and Optimal Transport}
Flow Matching~\cite{lipman2022flow} has recently emerged as a powerful alternative to diffusion models. The Conditional Flow Matching (CFM) framework has demonstrated superior performance and stability in deterministic generation tasks. Recent advancements have deeply explored its Optimal Transport (OT) properties; methods such as Rectified Flow~\cite{liu2022flow} and Optimal Transport Conditional Flow Matching (OT-CFM)~\cite{tong2023improving} significantly enhance sampling efficiency by linearizing probability paths.

Recent advancements have addressed the limitations of standard Gaussian priors by employing Gaussian Mixture Models (GMMs) as Conditional Prior Distributions (CPD)~\cite{issachar2025designing}. By minimizing the optimal transport cost, GMM priors substantially improve the modeling of complex data distributions. Building upon these insights, we pioneer the application of GMM-based flow matching to zero-shot vector sketch generation. Specifically, we embed a continuous GMM prior within the CLIP latent space. This approach effectively translates isolated, discrete textual categories into a continuous semantic distribution, thereby bridging the cross-modal gap between textual concepts and continuous visual trajectories.

\section{Method}
\begin{figure}[t]
    \centering
    \includegraphics[width=\linewidth]{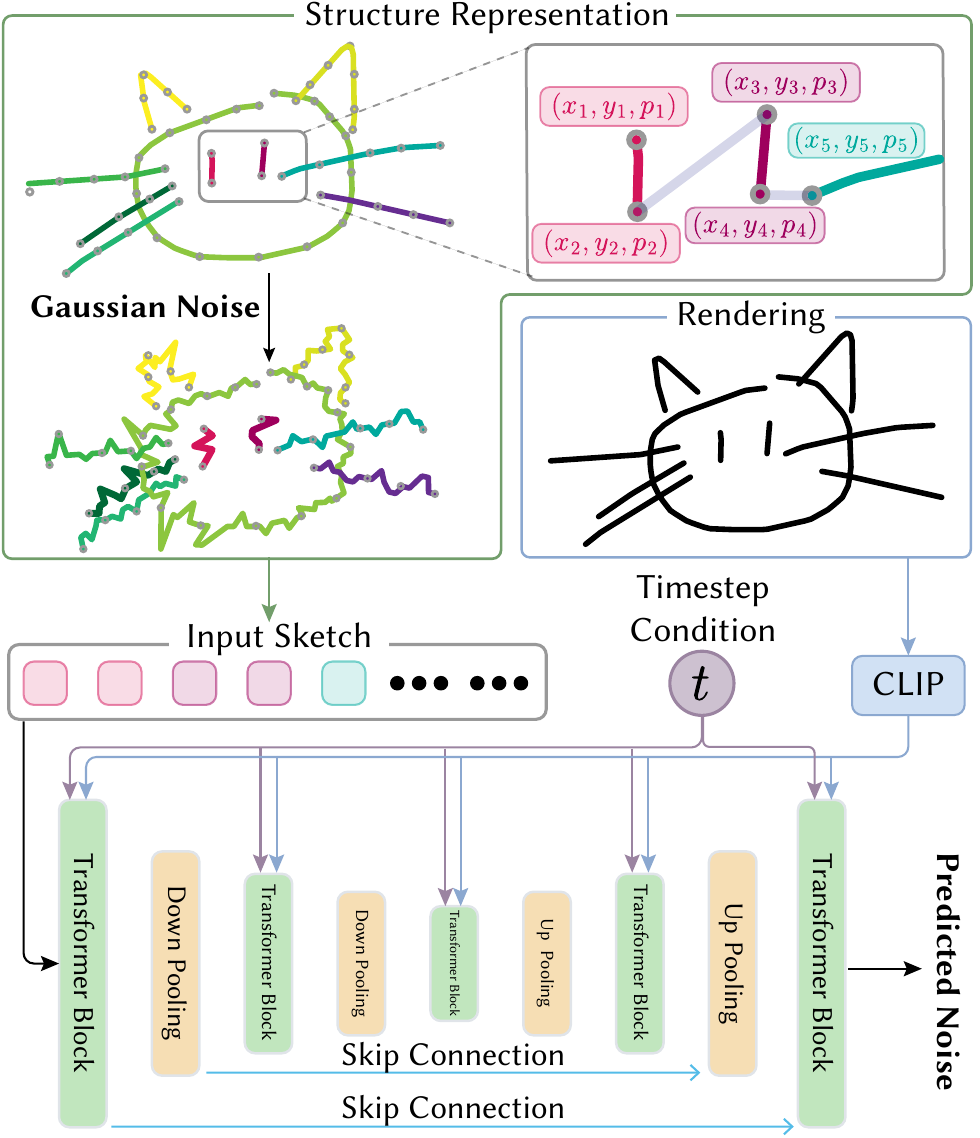}
    \caption{\textbf{Architecture of the Hybrid Diffusion Decoder.} The model employs an interleaved generative backbone combining U-Net-style bottlenecks for local geometric extraction with Transformer blocks for global dependency modeling. Condition embeddings (diffusion timestep $t$ and target CLIP embedding) are injected via AdaLN to predict the added noise.}
    \Description{Architecture diagram showing an ordered cat trajectory represented by coordinate and pen-state triples, its noisy version, and its rendering. Sequence tokens enter a multiscale encoder-decoder with interleaved Transformer blocks and skip connections; timestep and CLIP conditions are injected throughout, and the network predicts trajectory noise.}
    \label{fig:hybrid_decoder}
\end{figure}
\subsection{Sketch Representation and Hybrid Diffusion Decoder}
To better understand the required inputs for sketch synthesis, we first introduce the final generation module: the Hybrid Diffusion Decoder. As illustrated in the overall framework (Figure \ref{fig:pipeline_overview}), to decode target CLIP embeddings into vector stroke sequences, we design a conditional feature decoder. We represent the sketch trajectory as an equidistant sequence $S=\{(x_i,y_i,p_i)\}_{i=1}^L$ of $L$ points. Here, $(x_i,y_i)$ represents the 2D spatial coordinates. To facilitate a continuous diffusion process and avoid early over-commitment to discrete states at high-noise stages, we treat the pen-state indicator as a continuous variable $p_i \in \mathbb{R}$, scaled to a small variance range (e.g., $\{-0.1, 0.1\}$) rather than strictly binary. This ensures that $p_i$ specifies connectivity to the preceding point without interfering with the synthesis of the global geometric trajectory during the early generation steps.

For the generative backbone, we employ a Hybrid Diffusion Decoder (Figure \ref{fig:hybrid_decoder}) featuring an interleaved design of 1D U-Net convolutional blocks \cite{ronneberger2015u} and Transformer layers \cite{vaswani2017attention}. Within this architecture, the U-Net-style operations handle local geometric feature extraction and multi-resolution downsampling and upsampling, while the interleaved Transformer blocks process long-range global dependencies at their respective resolution scales.

For the conditioning mechanism, the diffusion timestep $t$ and the target rendered-sketch CLIP embedding (denoted as $z_{\text{target}}$) are independently processed via sinusoidal positional encodings and linear projections. These representations are then summed to form a unified condition vector, which is injected into the hidden features of each U-Net and Transformer block using Adaptive Layer Normalization (AdaLN). The decoder is trained using a standard denoising score matching objective \cite{song2020score, ho2020denoising}:
$$\mathcal{L}_{\text{Diff}}=\mathbb{E}_{x_0,\epsilon,t}\left[\left\|\epsilon-\epsilon_\theta(x_t,t,z_{\text{target}})\right\|^2\right]$$

We observe that, by using continuous CLIP embeddings rather than category labels as its semantic conditioning signal, the decoder can translate the visual semantics encoded in a rendered-sketch embedding into coherent strokes, including for unseen concepts. Therefore, the primary objective of our framework becomes acquiring the target sketch CLIP embedding directly from a textual prompt. We address this modality gap by constructing a continuous prior in the CLIP latent space, as detailed in the following section.

\subsection{GMM Prior Construction in CLIP Latent Space}
To address the scarcity of fine-grained text-sketch pairs, we formulate the cross-modal alignment as a distribution transport problem in the CLIP latent space. Given that available datasets predominantly provide categorical labels rather than rich textual descriptions, we leverage the pre-trained CLIP text encoder \cite{radford2021learning} to extract discrete text embeddings $z_{\text{text}}^k$ for each category $k$.

However, treating these isolated text embeddings as direct initial states does not provide continuous neighborhoods in which prompts outside the training vocabulary can be transported. To bridge this modality gap, we introduce a noise injection mechanism to construct a continuous Gaussian Mixture Model (GMM) prior. At initial state $t=0$, we define the prior distribution $p_0(z)$ as:
$$p_0(z)=\frac{1}{K}\sum_{k=1}^K\mathcal{N}(z;z_{\text{text}}^k,\sigma^2I)$$
This formulation transforms sparse concept anchors into a continuous probability density field, establishing a smooth semantic manifold from which our transport mapping originates. Here, $\sigma$ represents the standard deviation of the injected noise. We treat $\sigma$ as an adaptive parameter to dynamically modulate the prior, as elaborated in Sec.~\ref{sec:variance}.

\subsection{Cross-Modal Alignment via OT-Flow Matching}\label{sec:OT-Flow_Matching}
Given the constructed GMM prior $p_0(z)$ and the target sketch empirical distribution $p_1(z)$, our goal is to establish a deterministic mapping between these two modalities.

Conditional paradigms like unCLIP~\cite{ramesh2022hierarchical} assume a standard normal prior $p_0(z)=\mathcal{N}(0,I)$ and treat the text embedding as a condition. However, when trained on data limited to discrete categorical clusters, such formulations struggle to generalize. Ambiguous queries often cause the model to output a probabilistic mixture of isolated distributions, meaning it stochastically snaps to known modes rather than achieving true semantic interpolation.

To address this limitation, we directly formulate the continuous GMM as the base prior $p_0(z)$ and utilize Optimal Transport (OT) Conditional Flow Matching \cite{lipman2022flow, tong2023improving}. The generative process follows the ODE $\mathrm{d}z_t = v_t(z_t) \mathrm{d}t$. Crucially, rather than learning an arbitrary flow, we leverage an OT coupling between $p_0$ and $p_1$ to construct the target vector field. In practice, to maintain semantic consistency during training, we construct the coupling conditionally based on semantic categories. Specifically, given a category $k$, we independently sample the initial state $z_0 \sim \mathcal{N}(z_{\text{text}}^k, \sigma^2 I)$ ($\sigma$ is the adaptive prior variance detailed in Sec.~\ref{sec:variance}) and the target state $z_1 \sim p_1(z|k)$. The flow matching model is then trained to regress the optimal transport vector field, which trivially forms a straight path $z_0 \to z_1$, using the Conditional Flow Matching objective \cite{lipman2022flow}:
$$\mathcal{L}_{\text{CFM}}=\mathbb{E}_{t,p_0(z_0|k),p_1(z_1|k)}\left[\left\|v_\theta(z_t,t,c,\sigma)-(z_1-z_0)\right\|^2\right]$$
where the intermediate state is parameterized as $z_t=(1-t)z_0+tz_1$.

This OT formulation minimizes transport cost and yields straight-path, constant-velocity trajectories with two useful properties. (1) \textbf{Semantic Direction Preservation:} Straight paths preserve semantic feature directions in the CLIP latent space. (2) \textbf{Non-Crossing Sampling Paths:} The learned ODE flow produces non-intersecting paths that preserve the relative organization of semantic regions, allowing unseen queries to follow the surrounding transport toward semantically related regions of the rendered-sketch CLIP space.

\subsection{Adaptive Prior Variance Injection}\label{sec:variance}
Empirical results suggest that a single fixed noise scale $\sigma$ is sub-optimal for constructing the initial GMM prior across all semantic categories. Different semantic concepts intrinsically require varying degrees of spatial dispersion to achieve optimal generation fidelity.

To address this, we introduce a tunable prior variance mechanism. We condition the flow-matching vector field on the prior variance as $v_\theta(z_t,t,c,\sigma)$. Specifically, to construct this conditioning mechanism, the flow timestep $t$, the text embedding $c$, and the logarithm of the prior variance $\log(\sigma)$ are independently processed via sinusoidal positional encodings and linear projections. These representations are then summed to form a unified condition vector, which is injected into the hidden features of the flow matching network using Feature-wise Linear Modulation (FiLM) \cite{perez2018film}.

This allows the generated marginal distribution at the endpoint to be dynamically modulated, denoted as $\hat{p}_1(z\mid z_{\text{test}},\sigma)$. During training, we randomly sample standard deviations $\sigma$ to modulate the initial GMM prior, directly shaping the generative trajectory. During inference, this formulation allows users to adaptively adjust $\sigma$ based on specific prompts. Specifically, we introduce a scaling parameter $\gamma$ to control the sampling standard deviation of the query point relative to the standard deviation of the input condition.

\section{Experiments}\label{sec:experiments}

\subsection{Experimental Setup}

\paragraph{Datasets.} We use all 345 categories in the QuickDraw release distributed for Sketch-RNN. Each category contains 70,000 training, 2,500 validation, and 2,500 test sketches, yielding 24.15 million, 862,500, and 862,500 samples in the respective splits. We extract stroke trajectories for sequence-level modeling. Furthermore, we conducted experiments on cross-domain transfer datasets to evaluate the model's generalizability; these extensive results are provided in the supplementary material.

\paragraph{Implementation Details.} For data preprocessing, we resample sketch sequences to $L=256$ and scale binary pen-state indicators to $\{-0.1, 0.1\}$ to prevent premature commitment to discrete states during high-noise diffusion stages. During training, sketches are rendered in black and white for CLIP feature extraction via the pre-trained ViT-B-32. For visual presentation in this paper, generated stroke trajectories are explicitly color-coded from light to dark to illustrate the sequential drawing order. The OT-Flow Matching model is parameterized by a residual MLP with FiLM layers, while the Hybrid Decoder utilizes a 1D U-Net interleaved with Transformer blocks. The transport model and decoder are jointly optimized for 1,221,444 steps (518 epochs, using 20\% of the shuffled training loader per epoch) with AdamW and a total batch size of 256. Training takes approximately three days on eight NVIDIA RTX 3090 GPUs. At inference, we use 60 steps for both components: Heun's method integrates the transport ODE, followed by 60 DDPM denoising steps in the decoder. Comprehensive architectural and training hyperparameters are provided in the supplementary material.

\paragraph{Evaluation Metrics.} Visual fidelity is evaluated using FID on $256\times256$ rasterized renderings. Cross-modal semantic alignment is measured by CLIP Score using the prompt ``a sketch of [category]''.
To assess trajectory abstraction, we evaluate the \textit{RDP Point Count} by computing the absolute number of remaining anchor points after applying the Ramer-Douglas-Peucker algorithm \cite{ramer1972iterative}. Before simplification, each trajectory is isotropically normalized so that the longest side of its bounding box is 1,000; we then use a tolerance of $\epsilon=2.0$. Fewer remaining points indicate higher abstraction, avoiding biases from the original sketch resolution.

\paragraph{Baselines.} We compare representative generative paradigms: (1) \textbf{SketchRNN}~\cite{ha2017neural}, an LSTM-VAE trajectory model; (2) \textbf{StrokeFusion}~\cite{zhou2026strokefusion}, a latent trajectory diffusion model; (3) \textbf{NeuralSVG}~\cite{polaczek2025neuralsvg}, an SDS-based optimization approach; and (4) \textbf{SketchAgent}~\cite{vinker2025sketchagent}, an LLM-driven planning agent. To ensure fair comparison, we omit FID scores for optimization-based and LLM-driven methods, as they are not explicitly trained on our data distribution. For complementary open-prompt comparison, we additionally evaluate a two-stage pipeline combining T2I and CLIPasso~\cite{vinker2022clipasso} on matched unseen prompts; the complete qualitative comparison is provided in the supplementary material.

\begin{figure}[t]
    \centering
    \includegraphics[width=\linewidth]{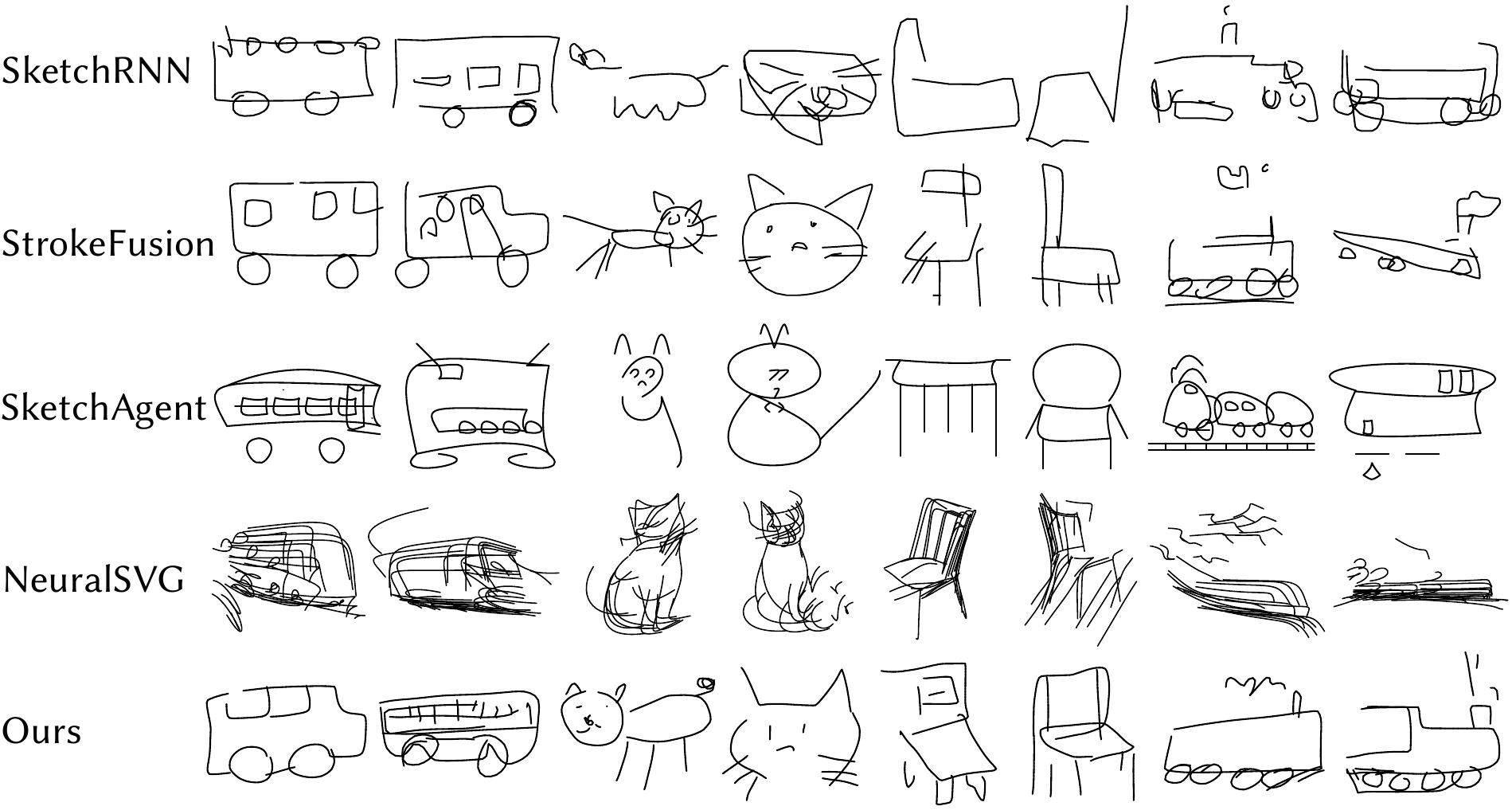}
    \caption{
    \textbf{Qualitative comparison on QuickDraw.}
    Generated samples for four categories (\textit{Bus}, \textit{Cat}, \textit{Chair}, and \textit{Train}) from baseline methods and ours.
    }
    \Description{A five-row comparison grid for SketchRNN, StrokeFusion, SketchAgent, NeuralSVG, and SketchFlow. Each row contains two generated examples of buses, cats, chairs, and trains; SketchFlow outputs use concise object-level strokes, while several baselines show missing parts, rigid geometry, or dense overlapping curves.}
    \label{fig:qual}
\end{figure}

\subsection{Comparison with Existing Methods}
To evaluate our approach against existing state-of-the-art methods, we select four representative categories: \textit{Bus}, \textit{Cat}, \textit{Chair}, and \textit{Train}. These categories exhibit relatively complex structural characteristics and are included in the known classes of the QuickDraw dataset, facilitating a fair and direct in-domain generation-quality comparison with other QuickDraw-based methods.

\begin{figure}[t]
    \centering
    \includegraphics[width=\linewidth]{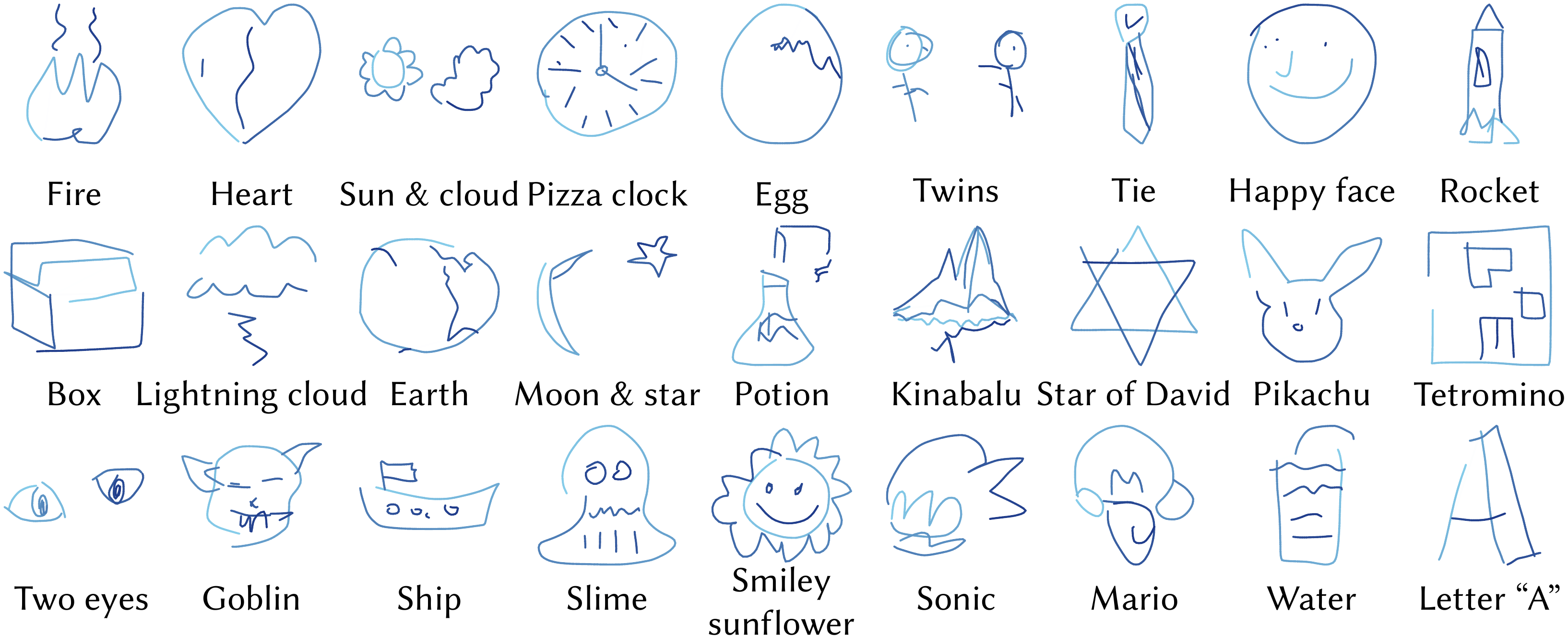}
    \caption{
    \textbf{Zero-shot generation beyond the training vocabulary.}
    All prompts shown here lie outside the QuickDraw taxonomy, yet the model produces coherent and style-consistent vector sketches.
    }
    \Description{Three-row gallery of blue vector sketches labeled with prompts outside the training vocabulary, including fire, heart, sun and cloud, pizza clock, twins, tie, rocket, Earth, moon and star, potion, Kinabalu, Pikachu, Tetromino, two eyes, Goblin, ship, slime, Sonic, Mario, water, and the letter A.}
    \label{fig:gallery}
\end{figure}

For each category, we generate 10,000 samples and compute the three evaluation metrics introduced in Sec.~\ref{sec:experiments}. To generate a specific category, the category name is used as the conditioning text input. Additional stochasticity is introduced by adding Gaussian noise scaled by the factor $\gamma$, as defined in Sec.~\ref{sec:variance}. We report our results under varying noise scales $\gamma \in \{0.6, 0.8, 1.0\}$, while the base prior variance $\sigma$ is fixed at $0.025$. The corresponding quantitative evaluations and qualitative visualizations are presented in Table~\ref{tab:quickdraw-results} and Figure~\ref{fig:qual}, respectively.

As shown in Table~\ref{tab:quickdraw-results}, our method achieves significant improvements in the FID metric compared to all learning-based baselines across all categories. This substantial margin demonstrates that our model possesses superior representation and feature learning capabilities. Although some baseline methods achieve lower RDP scores, this is primarily because they tend to learn and generate overly simplistic or rudimentary sketches, which is demonstrated by the qualitative results in Figure~\ref{fig:qual}.

Furthermore, our generated results achieve CLIP scores comparable to those of NeuralSVG and SketchAgent. However, the higher RDP scores of these baseline methods suggest a larger number of points after simplification, which often corresponds to cluttered and unstructured strokes. While such visual complexity may inadvertently inflate CLIP recognition scores, it deviates from actual human drawing behaviors. In contrast, qualitative comparisons in Figure~\ref{fig:qual} illustrate that \SysName{} better captures the abstract characteristics of hand-drawn sketches, producing trajectories that are concise, coherent, and effective at conveying target concepts.

\begin{table*}[t]
\centering
\caption{Quantitative results on QuickDraw \textit{Bus}, \textit{Cat}, \textit{Chair}, and \textit{Train}. Bold and underlined values indicate the best and second-best performances, respectively.}
\label{tab:quickdraw-results}
\begin{tabular}{lcccccccccccc}
\toprule
& \multicolumn{3}{c}{\textit{Bus}} 
& \multicolumn{3}{c}{\textit{Cat}} 
& \multicolumn{3}{c}{\textit{Chair}} 
& \multicolumn{3}{c}{\textit{Train}} \\
\cmidrule(lr){2-4} \cmidrule(lr){5-7} \cmidrule(lr){8-10} \cmidrule(lr){11-13}
Method & FID $\downarrow$ & CLIP $\uparrow$ & RDP $\downarrow$ & FID $\downarrow$ & CLIP $\uparrow$ & RDP $\downarrow$ & FID $\downarrow$ & CLIP $\uparrow$ & RDP $\downarrow$ & FID $\downarrow$ & CLIP $\uparrow$ & RDP $\downarrow$ \\
\midrule
SketchRNN                & 66.103 & 0.266 & \textbf{79.571}  & 70.704 & 0.210 & \textbf{68.591}  & 88.294 & 0.227 & \underline{25.351}  & 83.578 & 0.236 & \textbf{78.830} \\
StrokeFusion             & 59.329 & 0.278 & 86.690  & 52.940 & 0.260 & 85.730  & 33.463 & 0.277 & 36.460  & 59.418 & 0.247 & 102.910 \\
SketchAgent              &   --   & 0.258 & 124.429 &   --   & 0.214 & 93.684  &   --   & 0.267 & 40.200  &   --   & 0.234 & 98.400 \\
NeuralSVG                &   --   & 0.224 & 278.446 &   --   & \underline{0.261}& 267.091 &   --   & \textbf{0.304} & 225.663 &   --   & 0.225 & 256.168 \\
Gaussian-Prior& 9.556& 0.265& 98.53& 24.568& 0.148& 97.15& 17.572& 0.165& 38.45& 15.180& 0.183&99.06\\
Ours ($\gamma=0.60$)     & 19.381 & \textbf{0.287} & \underline{85.47}& 27.977 & \textbf{0.262} & \underline{79.77}& 41.123 & \underline{0.294} & \textbf{24.30}& 15.347 & \textbf{0.260} & \underline{92.29}\\
Ours ($\gamma=0.80$)     &  \underline{9.004} & \underline{0.284} & 92.12& \underline{17.899} & 0.261 & 86.33& \underline{21.545} & 0.287 & 29.75&  \underline{8.217} & \underline{0.256} & 96.21\\
Ours ($\gamma=1.00$)     &  \textbf{5.810} &  0.276 & 100.35& \textbf{14.423} & 0.254 & 95.75& \textbf{11.718} & 0.280 & 36.65&  \textbf{7.181} & 0.248 & 103.79\\
\rowcolor{gray!15}
Dataset& 0& 0.284& 91.88& 0& 0.259& 77.83& 0& 0.283& 29.29& 0& 0.266&101.89\\
\bottomrule
\end{tabular}
\end{table*}

\subsection{Zero-Shot Generation Beyond the Training Vocabulary}
Beyond generating known classes from the training dataset, our model demonstrates zero-shot generalization to labels outside the 345-class QuickDraw training vocabulary through OT-Flow Matching in the CLIP latent space. Figure~\ref{fig:gallery} presents results for several unseen prompts, with additional results provided in the supplementary material. These include concepts such as \textit{Pikachu}, \textit{Tetromino}, and \textit{Goblin}, as well as recognizable outputs for the composite and plural prompts \textit{Sun \& cloud} and \textit{Two eyes}. These results demonstrate that \SysName{} leverages visual semantics encoded by CLIP to generate vector sketches for unseen concepts.

\begin{table}[t]
    \centering
    \caption{Mean CLIP Score across 44 prompts absent from the QuickDraw training vocabulary. Each value averages 32 samples per prompt without best-of-$N$ selection ($\gamma=1.0$).}
    \label{tab:unseen-label-results}
    \setlength{\tabcolsep}{10pt}
    \begin{tabular}{lc}
        \toprule
        Method & Mean CLIP Score $\uparrow$ \\
        \midrule
        Gaussian Prior & 0.2024 \\
        Interp CLIP & 0.2034 \\
        \SysName{} (Ours) & \textbf{0.2480} \\
        \bottomrule
    \end{tabular}
\end{table}

To quantify generalization beyond selected examples, we evaluate all 44 prompts whose complete labels are absent from the 345-class training vocabulary. For each prompt, we generate 32 samples and report their mean CLIP Score as a label-alignment measure, without sample selection. The Interp CLIP baseline fits affine weights over the top-8 nearest QuickDraw text anchors, allowing extrapolation, and applies the weights to the corresponding sketch-latent anchors. As shown in Table~\ref{tab:unseen-label-results}, \SysName{} achieves 0.2480, compared with 0.2024 for the Gaussian-Prior baseline and 0.2034 for Interp CLIP, outperforming Interp CLIP on all 44 prompts. Complete per-prompt results are provided in the supplementary material.

Figure~\ref{fig:exp_complex_prompt} further shows that the model responds to semantic modifiers and actions, although it was trained exclusively on discrete category names. For prompts such as \textit{a running cat}, \textit{a jumping cat}, and \textit{a stretching cat}, the generated trajectories reflect distinct poses while maintaining the abstract hand-drawn style. This provides qualitative evidence that the learned CLIP-space transport retains modifier semantics beyond the category names used for training.

\begin{table}[t]
    \centering
    \caption{User study results comparing \SysName{} with \textbf{NeuralSVG} and \textbf{SketchAgent} across five metrics. The scores indicate the average percentage of user preference for each method.}
    \label{tab:user_study_results}
    \resizebox{\linewidth}{!}{
    \begin{tabular}{lccc}
        \toprule
        \textbf{Metric} & \SysName{} & \textbf{NeuralSVG} & \textbf{SketchAgent} \\
        \midrule
        Abstraction Fidelity & 72.85\% & 12.52\% & 14.63\% \\
        Semantic Alignment   & 56.75\% & 32.52\% & 10.73\% \\
        Human-likeness       & 70.73\% & 16.59\% & 12.68\% \\
        Stroke Rationality   & 61.79\% & 26.18\% & 12.03\% \\
        Aesthetic Quality    & 51.71\% & 37.24\% & 11.05\% \\
        \bottomrule
    \end{tabular}
  }
\end{table}

\begin{figure}[t]
    \centering
    \includegraphics[width=\linewidth]{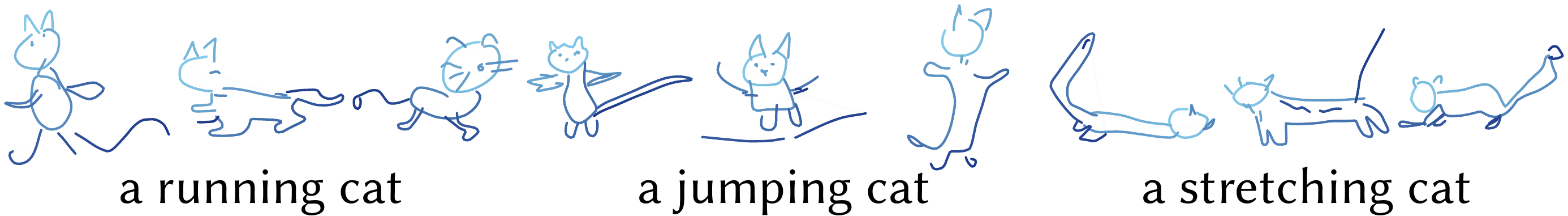}
    \caption{\textbf{Zero-Shot Generation with Semantic Modifiers.} Generation results for specific action prompts. Three distinct results are shown for each prompt.}
    \Description{Nine blue vector cat sketches arranged in three groups. Running cats have extended legs and horizontal motion, jumping cats have lifted or arched bodies, and stretching cats have elongated bodies and tails.}
    \label{fig:exp_complex_prompt}
\end{figure}

\begin{figure}[t!]
    \centering
    \includegraphics[width=\linewidth]{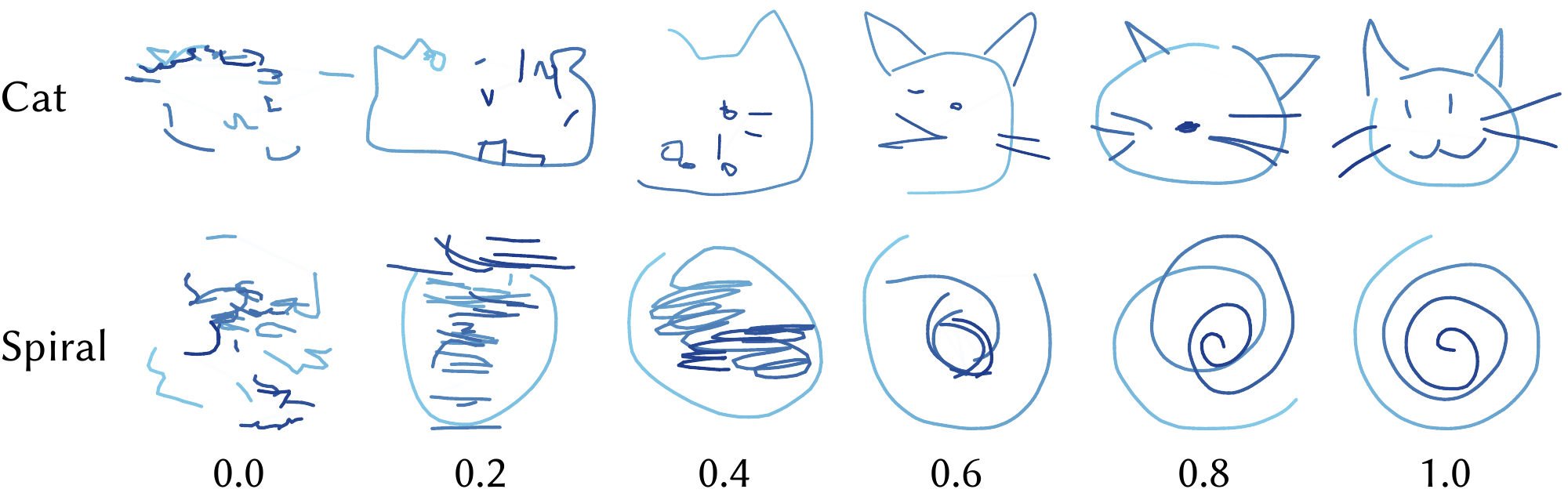}
    \caption{\textbf{Ablation of Flow Mapping.} Results of linear interpolating the condition from raw text embeddings ($\lambda=0.0$) to Flow-mapped embeddings ($\lambda=1.0$) at intervals of 0.2. The raw text embedding alone is insufficient to guide structured sketch synthesis.}
    \Description{Two rows show cat and spiral generations at interpolation values from 0.0 to 1.0. At low values the strokes are fragmented and disorganized; as the flow-mapped condition increases, they progressively form a recognizable cat face and a clean spiral.}
     \label{fig:exp_ablation_flow}
\end{figure}

\begin{figure}[t!]
    \centering
    \includegraphics[width=\linewidth]{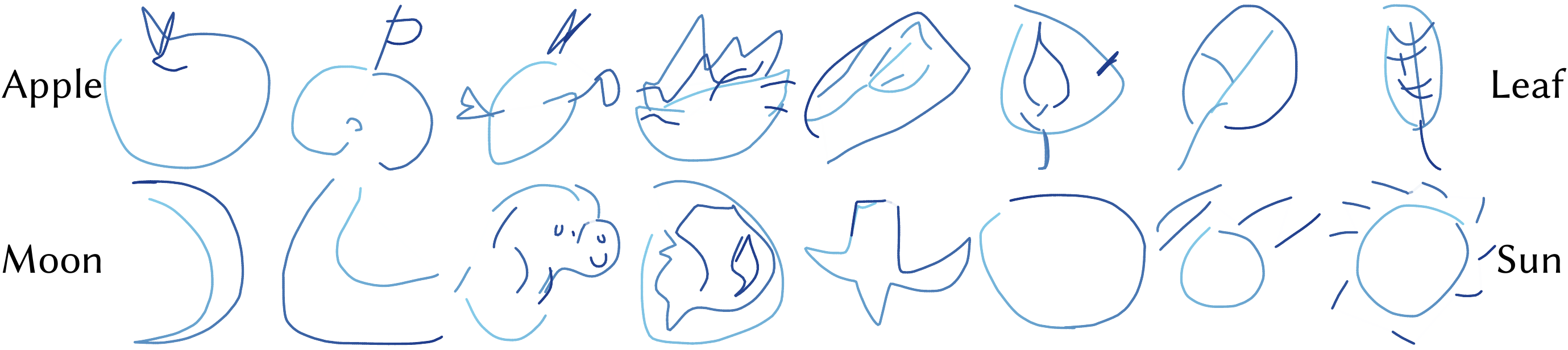}
    \caption{\textbf{Ablation of Gaussian-Prior baseline} Latent interpolation results produced by the baseline. Unlike  our approach, this baseline struggles to maintain structural coherence  during the transition. The intermediate sketches exhibit abrupt  topological shifts and a loss of visual semantics, highlighting the necessity of our GMM-based OT-Flow Matching design for establishing a continuous generative space. }
    \Description{Two interpolation rows produced by the Gaussian-prior baseline. The apple-to-leaf row passes through irregular mixed objects before reaching a leaf, and the moon-to-sun row contains abrupt shape changes and unrelated intermediate structures.}
     \label{fig:exp_ablation_unclip}
\end{figure}

\subsection{User Studies}\label{sec:User_studies}

Following~\cite{bhunia2022doodleformer, wang2024vq}, we conduct a user study of perceptual quality and human-likeness, comparing \SysName{} with NeuralSVG~\cite{polaczek2025neuralsvg} and SketchAgent~\cite{vinker2025sketchagent}. To account for the variance in generation quality among these methods and ensure a fair evaluation, we selected the single best result (highest CLIP score) from 10 generated samples per prompt. A total of 42 participants were recruited for this study. They assessed the sketches across five metrics: abstraction fidelity, semantic alignment, human-likeness, stroke rationality, and aesthetic quality. As shown in Table~\ref{tab:user_study_results}, \SysName{} consistently outperforms the comparison methods across all criteria. This strong preference demonstrates that our approach successfully captures the abstract semantic essence of concepts and naturally simulates authentic human drawing nuances (\eg, slight stroke variations).

\subsection{Continuous Interpolation} 
To demonstrate that our framework learns a continuous semantic manifold rather than merely overfitting to discrete training categories, we perform linear interpolation between distinct text embeddings. As shown in Figure~\ref{fig:exp_interpolation}, the model generates smooth and topologically coherent structural transitions between different concepts, validating the continuity of the established generative space.

\subsection{Ablation Studies}\label{sec:ablation}

CLIP text and sketch embeddings are misaligned; our OT-Flow Matching module bridges this cross-modal gap.
Figure~\ref{fig:exp_ablation_flow} ablates flow mapping by linearly interpolating the condition from the raw CLIP text embedding ($\lambda=0.0$) to its flow-mapped target ($\lambda=1.0$). Raw text conditioning misses the valid sketch manifold and produces unstructured, unrecognizable strokes. As the condition approaches the flow-mapped output, the trajectories smoothly self-organize into structurally coherent, recognizable sketches.
To empirically validate the necessity of our GMM prior formulation over the conventional conditional paradigm (Sec.~\ref{sec:OT-Flow_Matching}), we train a variant denoted as the ``Gaussian-Prior Baseline''. This model follows the unCLIP approach by assuming a standard normal prior $p_0(z)=\mathcal{N}(0, I)$ and treating the text embedding strictly as a condition.

 Table~\ref{tab:quickdraw-results} records the quantitative evaluation of this baseline, which exhibits a clear performance degradation compared to our method. This decline stems from the unCLIP paradigm's inability to explicitly align or organize the cross-modal distributions. Consequently, a perturbed conditional embedding from one category can easily drift into the latent representation of another, resulting in a noticeable probability of generating sketches that deviate from the input class.

\begin{figure*}[t]
    \centering
    \includegraphics[width=0.97\linewidth]{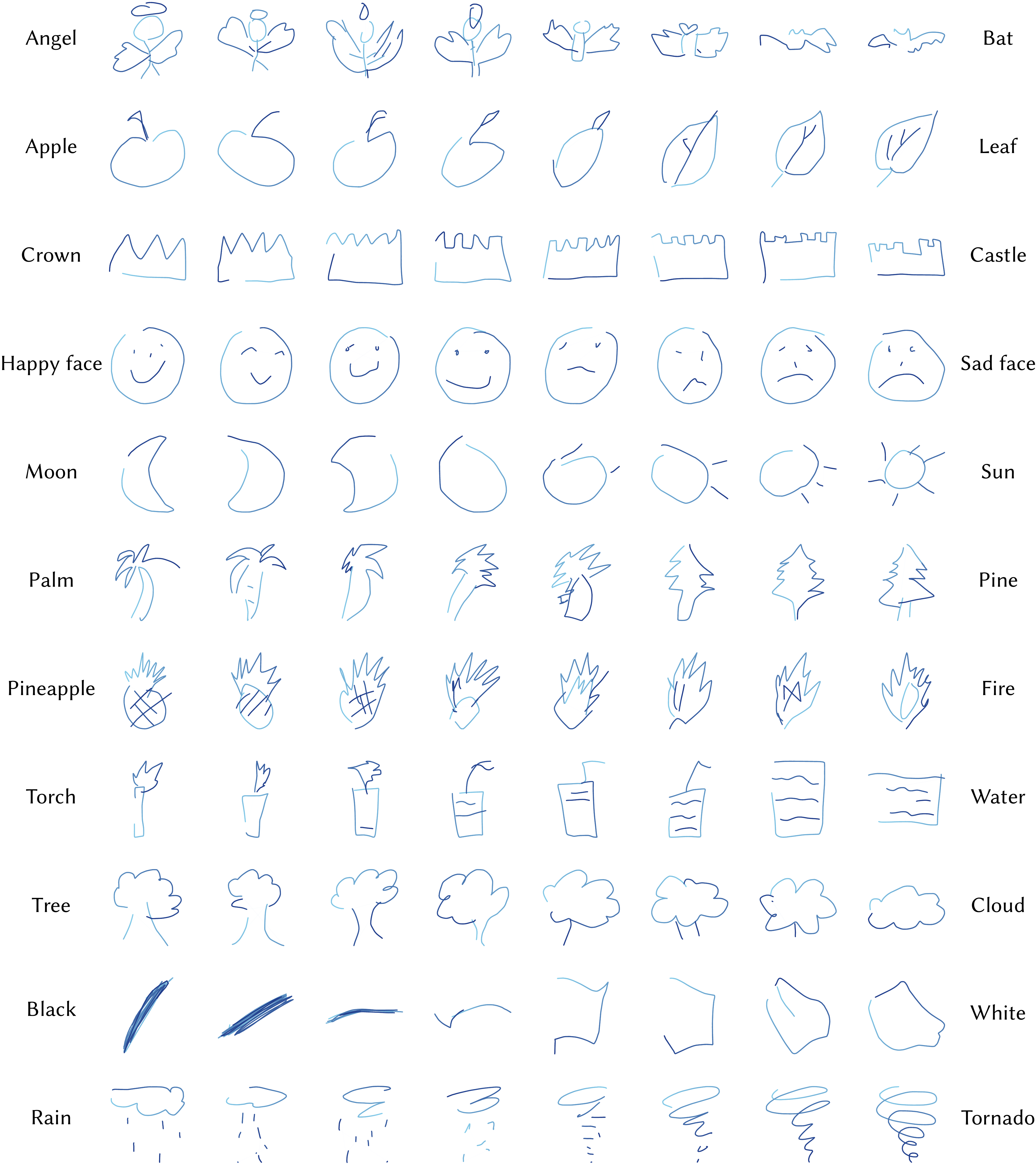}
    \caption{\textbf{Semantic Interpolation in CLIP Space.} Each row visualizes a linear interpolation path between the discrete text embeddings of two distinct concepts (e.g., smoothly morphing from an ``Apple'' to a ``Leaf'', or from a ``Palm'' to a ``Pine''). The intermediate generations maintain exceptionally uniform structural transitions and logical stroke topologies without disjointed visual jumps. This progressive morphing provides compelling evidence that our model successfully establishes a continuous generative manifold, rather than stochastically snapping to isolated known modes.}
    \Description{Nine-stage interpolation rows transform angel to bat, apple to leaf, crown to castle, happy face to sad face, moon to sun, palm to pine, pineapple to fire, torch to water, tree to cloud, black to white, and rain to tornado. Adjacent sketches change gradually in outline and internal strokes.}
    \label{fig:exp_interpolation}
\end{figure*}

More importantly, as discussed in Section~\ref{sec:OT-Flow_Matching}, this purely conditional model generalizes poorly to the unseen prompts considered in our experiments. To illustrate this, we replicate the latent interpolation experiment using the unCLIP baseline, with results shown in Figure~\ref{fig:exp_ablation_unclip} and Figure~\ref{fig:exp_interpolation}. The generated sketches exhibit abrupt structural shifts and rapidly lose visual readability. Rather than traversing a continuous semantic manifold, the model simply snaps between isolated training modes, failing to synthesize structurally coherent intermediate concepts.

\subsection{Image-Conditioned Generation}
\begin{figure*}[t]
    \centering
    \includegraphics[width=\linewidth]{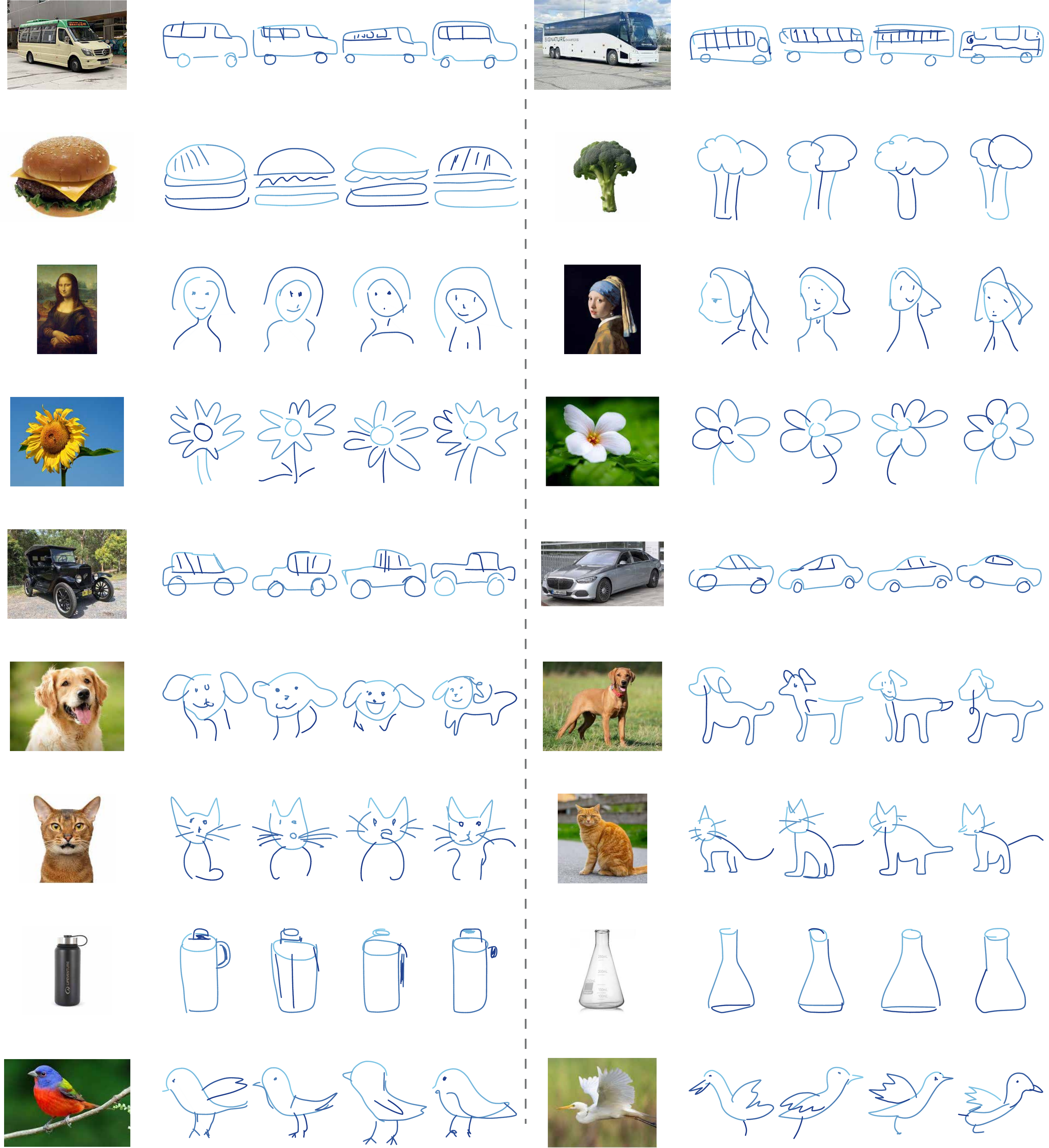}
    \caption{\textbf{Zero-shot Image-Conditioned Vector Sketch Synthesis.} As an intriguing extension, our model can seamlessly accept CLIP visual embeddings in place of textual prompts. This substitution enables a zero-shot stylized redrawing effect, where the generated vector sketches effectively capture the core semantic features and structural characteristics of the reference images.}
    \Description{Reference photographs are paired with four generated vector sketches each. The examples cover buses, foods, portraits, flowers, old and modern cars, dogs, cats, drink containers, laboratory glassware, and birds; the outputs preserve distinguishing silhouettes and parts while simplifying appearance into blue strokes.}
    \label{fig:image2sketch}
\end{figure*}

Because CLIP aligns vision and language in latent space, replacing the text prompt with a reference-image embedding extends \SysName{} to zero-shot semantic abstraction and style transfer without architectural changes or paired image-sketch fine-tuning. As shown in Figure~\ref{fig:image2sketch}, \SysName{} captures the input's core semantics, geometry, and pose in abstraction style.

\begin{figure}[t]
    \centering
    \includegraphics[width=\columnwidth]{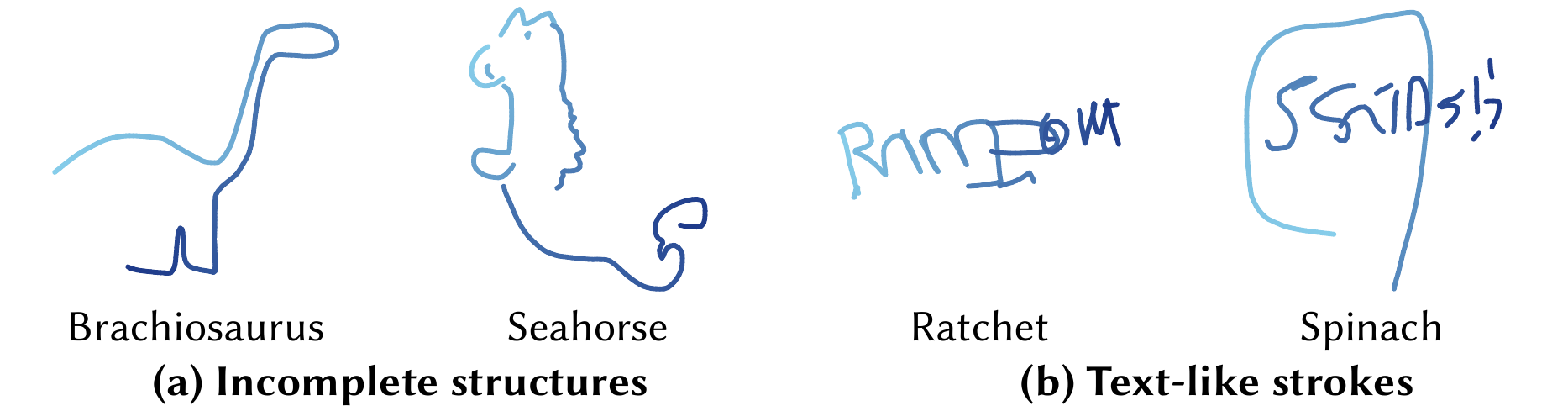}
    \caption{\textbf{Representative failure cases.} (a) The outputs remain recognizable as the prompted concepts but contain incomplete structures. (b) The generated trajectories form text-like strokes resembling the prompt words instead of the intended objects.}
    \Description{Four generated sketches showing incomplete structures for a brachiosaurus and a seahorse, followed by text-like outputs for the prompts ratchet and spinach.}
    \label{fig:failure_cases}
\end{figure}

\section{Conclusion \& Limitations}\label{sec:conclusion}

\paragraph{Conclusion.} We present \SysName{}, a zero-shot text-to-vector sketch framework that bridges discrete text concepts and continuous sketch features through OT-CFM and a continuous GMM prior in CLIP space. Its fast forward-pass generator achieves high visual quality, follows human drawing styles, and shows promising local zero-shot generalization beyond the QuickDraw vocabulary.

The results demonstrate that Optimal Transport geometry enables CLIP-mediated zero-shot transport under label-only textual supervision, reducing reliance on exhaustive fine-grained textual annotations for individual training sketches.

\paragraph{Limitations.} The text-to-sketch mapping remains imperfect. As shown in Figure~\ref{fig:failure_cases}, some outputs for prompts beyond the training vocabulary retain recognizable concept-level structure but omit parts of the intended shape, while others drift toward text-like strokes resembling the prompt word rather than depicting the intended object. Moreover, some unseen categories require increased search variance ($\gamma$) and still yield inconsistent results. We hypothesize that this limitation reflects the locally uniform Gaussian assumption around text anchors, whereas semantic manifolds can be complex and non-Gaussian. Future work will develop more precise latent mappings to better capture these non-Gaussian semantic structures.

\begin{acks}
We thank the anonymous reviewers for their constructive feedback and the user study participants for their time and effort. This work was supported in part by National Natural Science Foundation of China (62472287), Natural Science Foundation of Shenzhen City (JCYJ20250604181519025), and Scientific Development Fund from Guangdong Provincial Key Laboratory of Visual Media and Multidimensional Intelligence.
\end{acks}

\bibliographystyle{ACM-Reference-Format}
\bibliography{main}

\clearpage
\appendix



\section{Overview}

This supplementary material complements the main paper with the following content: (i) \textbf{preprocessing details}, including length-proportional point allocation, short-stroke handling, and final resampling; (ii) \textbf{implementation details} of \SysName{}, covering network architectures, training setup, and inference configurations with prompt examples and hyperparameter guidance; (iii) a \textbf{cross-domain transfer evaluation} on TU-Berlin and Creative-Mix with both quantitative and qualitative comparisons; (iv) an \textbf{extended zero-shot generation} gallery together with a broader analysis of NeuralSVG under varying stroke counts; and (v) \textbf{user study details}, including the questionnaire design, drawing-order visualization, and extended qualitative analysis.

\section{Preprocessing Details}

We convert each vector sketch to a fixed sequence of $n=256$ points, yielding a uniform drawing-time parameterization for the decoder. We next describe proportional point allocation and short-stroke handling.

\subsection{Length-Proportional Point Allocation}

Given a vector sketch containing $M$ strokes ($\mathbf{s}_1, \ldots, \mathbf{s}_M$), each with an arc length $L_k$, we employ a length-proportional strategy to distribute the total $n=256$ points:

\begin{enumerate}[leftmargin=*, nosep]
    \item \textbf{Initial Assignment:} Every stroke $\mathbf{s}_k$ is initially allocated one point.
    \item \textbf{Proportional Distribution:} The remaining $N' = n - M$ points are distributed among the strokes proportionally to their arc lengths $L_k$. The initial allocated point count $N_k^{\text{init}}$ for stroke $\mathbf{s}_k$ is:
    
$$ N_k^{\text{init}} = 1 + \text{round}\left(N' \cdot \frac{L_k}{\sum_{j=1}^{M} L_j}\right) $$

where the total points $\sum N_k^{\text{init}}$ is corrected to equal $n$ if rounding leads to a slight discrepancy.
\end{enumerate}

\subsection{Handling of Short Strokes}

In rare cases, especially for extremely short strokes, the initial allocation may result in a stroke being assigned $N_k^{\text{init}}=1$ point. To facilitate accurate, uniform resampling (which requires at least two points to define a segment) and better capture the trajectory, we perform a point re-allocation:

\begin{enumerate}[leftmargin=*, nosep]
    \item We identify any stroke $\mathbf{s}_k$ for which the allocated point count $N_k^{\text{init}}$ is exactly $1$.
    \item For each such stroke, we borrow one point from the stroke $\mathbf{s}_{\max}$ that currently holds the largest arc length $L_{\max}$ (and $N_{\max} \ge 2$ points). This re-allocation is performed until all strokes have $N_k \ge 2$ or the longest stroke can no longer afford the transfer, ensuring the minimal representation for a trajectory stroke is two points whenever possible.
\end{enumerate}

\subsection{Final Resampling and Encoding}

After the final point counts $N_k$ are determined, each stroke $\mathbf{s}_k$ is uniformly resampled along its arc length. The segment length for resampling is calculated as $l_k = L_k / (N_k - 1)$. The resulting points from all $M$ strokes are concatenated in their original drawing order to form the sequence $\mathbf{s} \in \mathbb{R}^{n \times 2}$. The final input $\tilde{\mathbf{s}} \in \mathbb{R}^{n \times 3}$, incorporating the pen-state flag $p_i$, is then constructed as detailed in the main paper. Regarding stroke preprocessing, we scale the binary pen-state indicators to $\{-0.1, 0.1\}$ and treat them as continuous variables. Empirically, we observe that this small variance scaling prevents the diffusion model from prematurely committing to discrete state transitions at early, high-noise generation stages. It effectively encourages the model to prioritize synthesizing the global geometric trajectory before resolving local stroke breaks.

\section{Implementation Details}
\label{sec:supp_impl_details}

\subsection{Network Architectures} For the OT-Flow Matching Model, the velocity field network $v_\theta$ is parameterized by a residual MLP equipped with Feature-wise Linear Modulation (FiLM) layers, featuring a hidden width of 1024 and a depth of 10 layers. During training, we apply an L2 regularization penalty on the predicted velocity vectors to minimize the kinetic energy of the flow. For the Hybrid Diffusion Decoder, the 1D U-Net backbone utilizes a base dimension of 128, with channel multipliers set to (1, 1.5, 2, 4). At each resolution scale, we interleave 2 Transformer blocks, where the number of attention heads increases progressively with the network depth (4, 4, 8, 8).

\subsection{Training Setup} The OT-Flow Matching model and Hybrid Diffusion Decoder are jointly trained for 1,221,444 optimizer steps (518 epochs, using 20\% of the shuffled training loader per epoch). Training takes approximately three days on 8 $\times$ NVIDIA RTX 3090 GPUs. We use AdamW with a learning rate of 1e-4, a weight decay of 1e-3, and $\beta$ values of (0.9, 0.999). The total batch size is 256. For the adaptive prior variance injection, we set the mean standard deviation to 0.025 and introduce a log-normal perturbation by randomly scaling it with a logarithmic coefficient sampled with a standard deviation of 0.25.

\subsection{Inference Configuration}

For inference, Heun's method integrates the transport ODE for 60 steps, and the Hybrid Diffusion Decoder then uses 60 DDPM denoising steps. In this section, we provide detailed inference settings and the specific text prompts used to generate the results showcased in the paper, particularly in the teaser figure.

\textbf{Prompts for Typographic and Symbolic Sketches.} For the stylized letters and numbers demonstrated in the teaser, we utilized specific descriptive prompts to guide the structure and style. Examples include:
{\setlength{\emergencystretch}{1em}
\begin{itemize}[leftmargin=20pt, nosep]
    \item ``A minimalist logo of the letter G with double-stroke effect'' 
    \item ``A simple letter s'' 
    \item ``A symbol of the number 2 with double-stroke effect''
\end{itemize}
}
Similar prompts were employed to synthesize the remaining typographic results.

\textbf{Prompts for Architectural Sketches.} For the complex building structures shown in the teaser, we applied the following detailed prompts to ensure structural coherence and specific architectural features:
\begin{itemize}[leftmargin=20pt, nosep]
    \item ``Twin towers.'' 
    \item ``A high tower with a spherical top and a sharp needle spire.'' 
    \item ``A building contains the copper onion-domed clock tower, and the flanking copper-domed minarets.'' 
    \item ``A church building with a central bell turret with a cross, arched three entrance portals.'' 
    \item ``A mosque featuring a large central dome and multiple smaller flanking domes with a prominent central minaret spire.''
\end{itemize}
All other explicitly annotated results in the manuscript were generated using their corresponding provided text prompts.

\textbf{Hyperparameter Settings.} The generation process is highly dependent on two key hyperparameters: the base prior variance $\sigma$ and the sampling noise scaling factor $\gamma$. We empirically established the following default configurations for different generation scenarios:
\begin{itemize} [leftmargin=20pt, nosep]
\item \textbf{Known categories:} $\sigma = 0.025$, $\gamma = 1.0$ 
\item \textbf{Prompts beyond the training vocabulary:} $\sigma = 0.025$, $\gamma = 0.4$
\item \textbf{Specific action prompts:} $\sigma = 0.04$, $\gamma = 0.4$ 
\item \textbf{Image-conditioned generation:} $\sigma = 0.04$, $\gamma = 0.0$
\end{itemize}

\textbf{Guidance on Parameter Tuning.} We use simple heuristics to adjust $\sigma$ and $\gamma$. The base prior variance $\sigma$ controls the semantic search space. If outputs are low quality and overly similar, we increase $\sigma$; if they vary excessively or lack structural consistency, we decrease it. The scaling factor $\gamma$ controls the trade-off between stability and diversity. We decrease $\gamma$ when generation requires high fidelity and strict adherence to the input condition.

\section{Cross-Domain Transfer Evaluation}

To comprehensively assess the model's cross-domain transfer ability beyond the primary QuickDraw training corpus, we further evaluate \SysName{} on three supplementary datasets:

\begin{itemize}[leftmargin=20pt, nosep]
    \item \textbf{TU-Berlin Sketch Dataset}: This benchmark contains 250 object categories and roughly 20,000 freehand black-and-white sketches, with approximately 80 sketches per class.
    \item \textbf{Creative Birds} (8,067 sketches) and \textbf{Creative Creatures} (9,097 sketches): These subsets of the ``Creative Sketch Generation'' dataset~\cite{ge2020doodlegen} contain imaginative sketches with part-level annotations. We merge them into a single evaluation dataset, denoted as \textbf{Creative-Mix}.
\end{itemize}

\begin{table}[t]
    \centering
    \caption{\textbf{Quantitative comparison of FID on Creative datasets.} Lower scores indicate better visual fidelity and structural similarity to the target distribution.}
    \label{tab:doodleformer_fid}
    \resizebox{\linewidth}{!}{
    \begin{tabular}{lccc}
        \toprule
        Method & Creative Birds & Creative Creatures & Creative Mix \\
        \midrule
        Doodleformer & 58.118 & 63.212 & 58.219 \\
        Ours & \textbf{34.444} & \textbf{32.938} & \textbf{32.181} \\
        \bottomrule
    \end{tabular}
    }
\end{table}

\textbf{Quantitative Results.} We conduct quantitative comparisons on the creative datasets, evaluating visual fidelity using the Fréchet Inception Distance (FID) metric. As reported in Table~\ref{tab:doodleformer_fid}, our method substantially outperforms the baseline Doodleformer across all data subsets. This remarkable reduction in FID indicates that our approach is highly capable of synthesizing sketches with superior visual quality and closer distributional alignment to the target domain, even in highly imaginative and abstract contexts.

\begin{figure}[t]
    \centering
    \includegraphics[width=\linewidth]{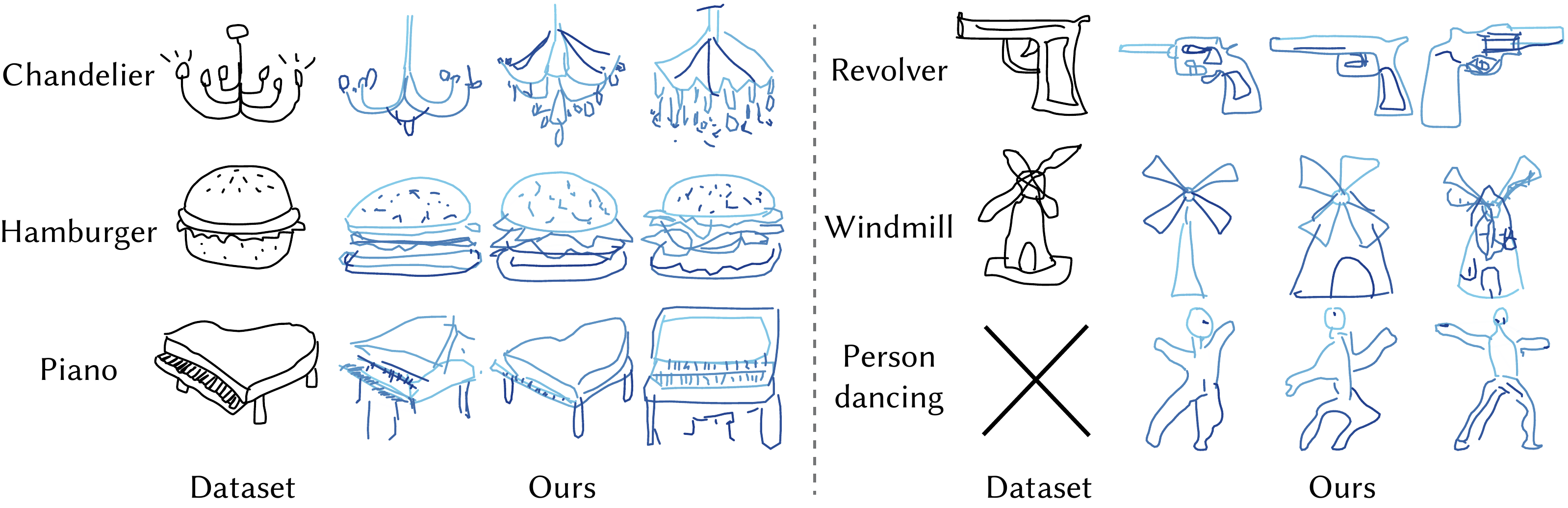}
    \caption{\textbf{Qualitative results on the TU-Berlin dataset.} Our model effectively learns the dataset's unique freehand style and produces coherent sketches, even for categories absent from the original dataset.}
    \Description{Dataset references and three SketchFlow samples are shown for chandelier, hamburger, piano, revolver, windmill, and person dancing. For person dancing, the dataset reference is marked unavailable while the generated figures show several dynamic poses.}
    \label{fig:tuberlin}
\end{figure}

\begin{figure}[t]
    \centering
    \includegraphics[width=\linewidth]{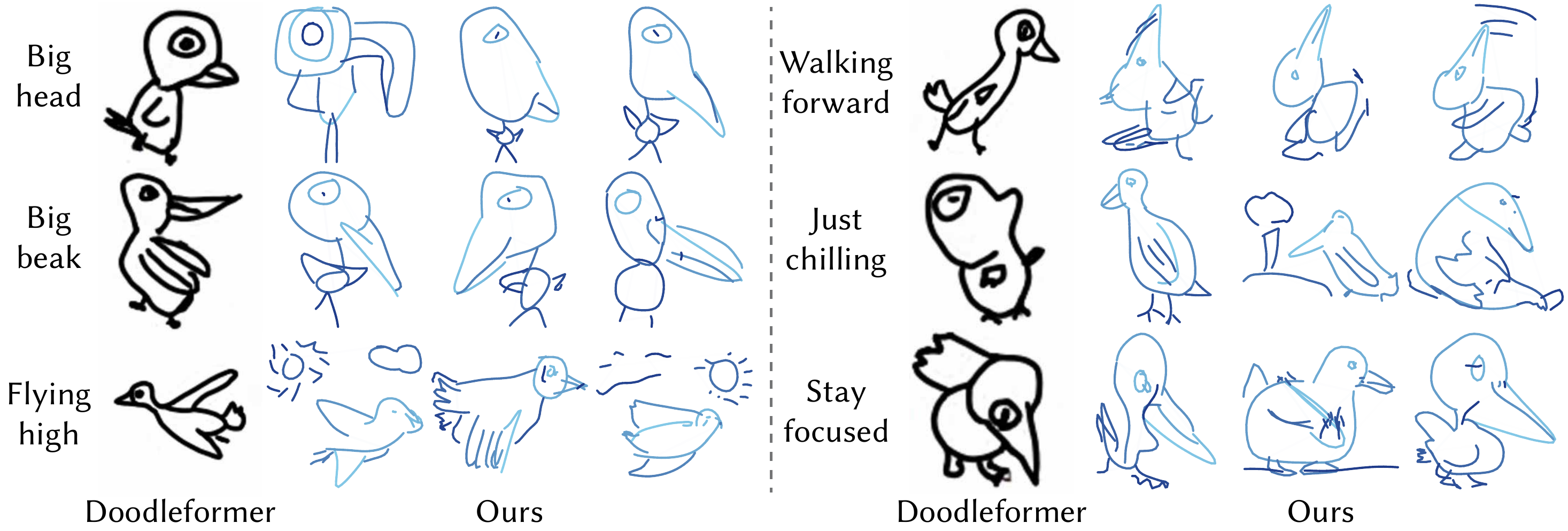}
    \caption{\textbf{Qualitative results on the Creative-Mix dataset.} Our method successfully adapts to the highly creative style while maintaining strict semantic control, visibly outperforming Doodleformer in overall structure and clarity.}
    \Description{Six creative bird prompts are compared between a Doodleformer reference and three SketchFlow samples: big head, big beak, flying high, walking forward, just chilling, and stay focused. Generated birds vary in pose while retaining the requested exaggerated trait or action.}
    \label{fig:creative}
\end{figure}

\textbf{Qualitative Results.} Qualitative visualizations further validate our model's strong transferability and stylistic adaptability. Figure~\ref{fig:tuberlin} illustrates the generation results on the TU-Berlin dataset. \SysName{} successfully captures the dataset's distinctive drawing style and generates structurally coherent sketches, including zero-shot examples for categories absent from the primary training corpus. Figure~\ref{fig:creative} compares results on Creative-Mix. Our method adapts to this highly stylized, imaginative dataset while maintaining semantic control over the target concepts, surpassing Doodleformer in both structural clarity and logical stroke composition.

\begin{table*}[t]
\centering
\caption{Additional quantitative comparisons on the QuickDraw dataset (\textit{Bus}, \textit{Cat}, \textit{Chair}, and \textit{Train}), specifically evaluating NeuralSVG with varying stroke counts. Bold and underlined values indicate the best and second-best performance, respectively.}
\label{tab:more-quickdraw-results}
\begin{tabular}{lcccccccc}
\toprule
& \multicolumn{2}{c}{\textit{Bus}} 
& \multicolumn{2}{c}{\textit{Cat}} 
& \multicolumn{2}{c}{\textit{Chair}} 
& \multicolumn{2}{c}{\textit{Train}} \\
\cmidrule(lr){2-3} \cmidrule(lr){4-5} \cmidrule(lr){6-7} \cmidrule(lr){8-9}
Method & CLIP$\uparrow$ & RDP$\downarrow$ & CLIP$\uparrow$ & RDP$\downarrow$ & CLIP$\uparrow$ & RDP$\downarrow$ & CLIP$\uparrow$ & RDP$\downarrow$ \\
\midrule
NeuralSVG-4              & 0.208 & \textbf{36.891}  & 0.212 & \textbf{36.374}  & 0.247 & 29.980  & 0.213 & \textbf{33.941} \\
NeuralSVG-8              & 0.215 & \underline{72.218}  & 0.228 & \underline{70.475}  & 0.271 & 58.554  & 0.215 & \underline{66.099} \\
NeuralSVG-16             & 0.219 & 142.376 & 0.248 & 137.232 & \underline{0.291} & 115.030 & 0.221 & 129.941 \\
NeuralSVG-32             & 0.224 & 278.446 & \textbf{0.261} & 267.091 & \textbf{0.304} & 225.663 & 0.225 & 256.168 \\
Ours ($\gamma$=0.60)     & \textbf{0.283} & 86.310  & \textbf{0.261} & 83.720  & 0.290 & \textbf{21.240}  & \textbf{0.259} & 87.140 \\
Ours ($\gamma$=0.80)     & \underline{0.279} & 89.080  & 0.255 & 87.980  & 0.282 & \underline{27.220}  & \underline{0.253} & 91.700 \\
Ours ($\gamma$=1.00)     & 0.270 & 94.370  & 0.248 & 89.090  & 0.276 & 36.360  & 0.245 & 97.690 \\
\bottomrule
\end{tabular}
\end{table*}

\section{Extended Zero-Shot Generation Beyond the Training Vocabulary}
\label{supp:subsec:ood_samples}
\begin{figure*}[t]
    \centering
    \includegraphics[width=0.95\linewidth]{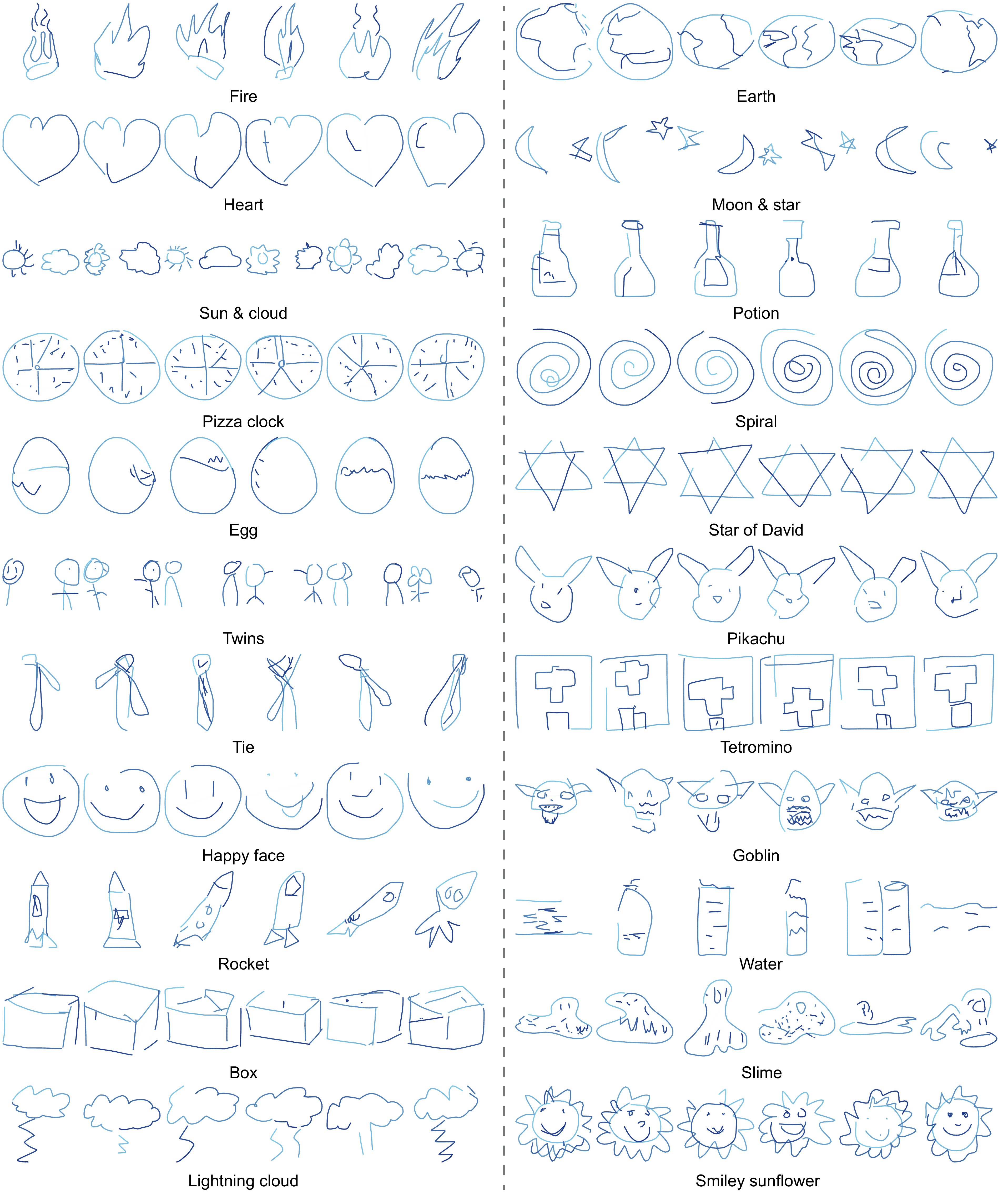}
    \caption{Extended zero-shot sketch generation beyond the training vocabulary. We show clear, visually appealing sketches for multiple prompts, with samples demonstrating generation diversity and a recurring hand-drawn style across random seeds.}
    \Description{Two-column gallery with six generated samples for each labeled prompt. The left column includes fire, heart, sun and cloud, pizza clock, egg, twins, tie, happy face, rocket, box, and lightning cloud; the right includes Earth, moon and star, potion, spiral, Star of David, Pikachu, Tetromino, Goblin, water, slime, and smiley sunflower.}
    \label{fig:extended_ood}
\end{figure*}

In the main paper, we showed a small number of examples generated from prompts outside the QuickDraw training vocabulary. Figure~\ref{fig:extended_ood} provides additional results for this setting. The model produces clear sketches for a range of unseen concepts, while multiple samples per prompt demonstrate generation diversity and a recurring hand-drawn style across random seeds. These results demonstrate CLIP-mediated zero-shot generation on the evaluated prompts.

\subsection{Unseen-Label Quantitative Evaluation}

Table~\ref{tab:unseen-label-complete} reports the complete per-prompt results for the 44-label evaluation summarized in the main paper. For each prompt, we report the nearest QuickDraw text anchor and its cosine similarity, together with the mean CLIP Score over 32 samples generated without sample selection.

\begin{figure}[t]
    \centering
    \includegraphics[width=\linewidth]{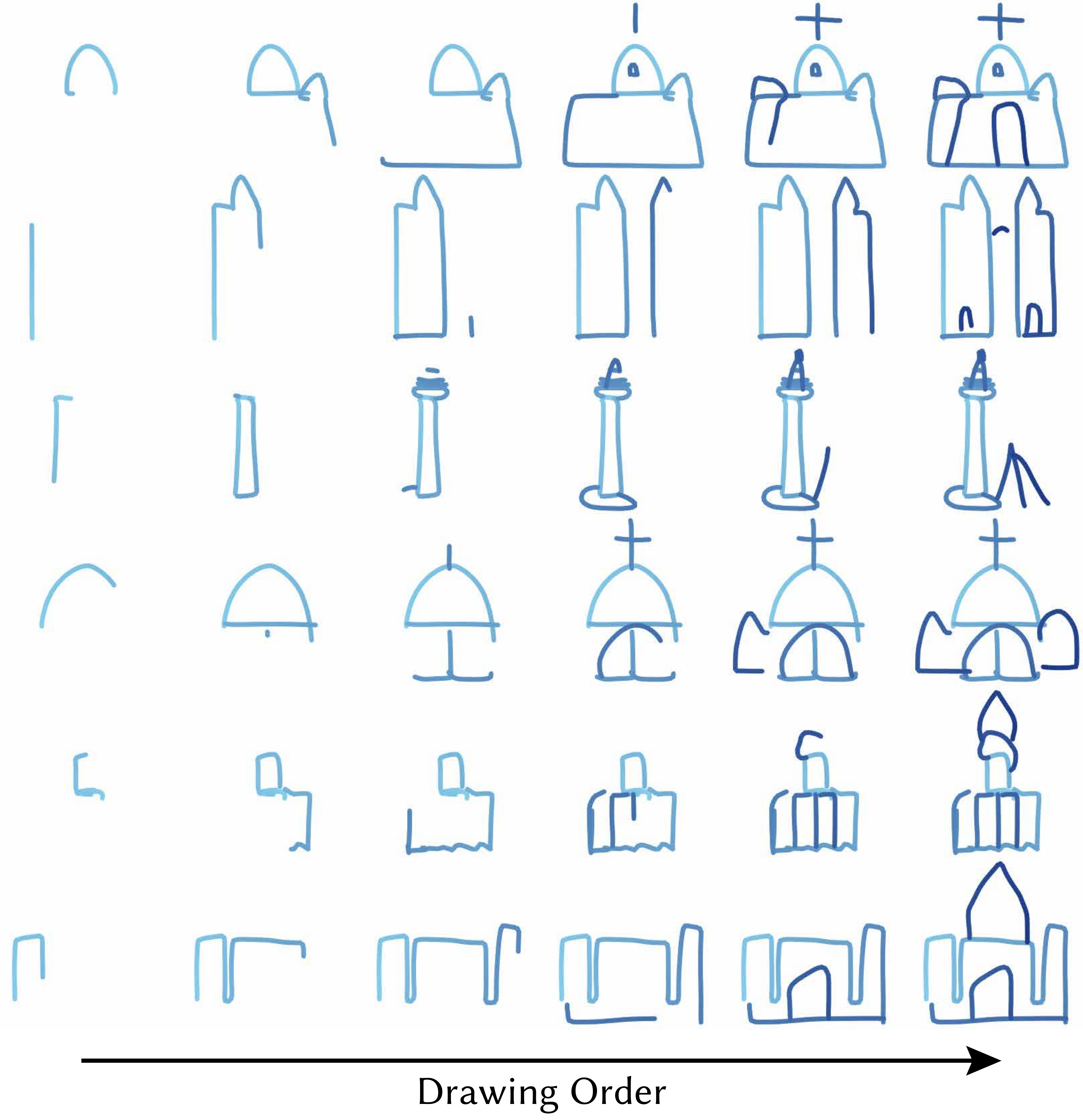}
    \caption{\textbf{Drawing Order Visualization.} To clearly illustrate the sequential nature of our generated vector sketches, stroke trajectories are color-coded according to their generation sequence, progressing from light blue (initial strokes) to dark blue (final strokes). Note that all generated sketches presented in this paper adopt this color-coding scheme to indicate the underlying drawing progression.}
    \Description{Five rows show six cumulative drawing stages for Malaysian landmark sketches. Each sequence begins with a few light-blue structural strokes and gains darker later strokes until the mosque, towers, and related building silhouettes are complete.}
    \label{fig:exp_draw_sequence}
\end{figure}

\begin{table*}[t]
\centering
\caption{\textbf{Complete unseen-label quantitative evaluation.} CLIP Score is used as a label-alignment proxy and averaged over 32 samples per prompt without sample selection. The nearest anchor and cosine similarity are computed in CLIP text space over the 345 QuickDraw training labels. Prompts are arranged in two side-by-side blocks for compact presentation. Bold indicates the best result among the three evaluated methods.}
\label{tab:unseen-label-complete}
{
\footnotesize
\setlength{\tabcolsep}{2.2pt}
\renewcommand{\arraystretch}{1.06}
\begin{tabular*}{\textwidth}{@{\extracolsep{\fill}}llrrrr@{\hspace{1.4em}}llrrrr@{}}
\toprule
\multicolumn{3}{c}{Prompt and nearest anchor} & \multicolumn{3}{c}{Mean CLIP Score}
& \multicolumn{3}{c}{Prompt and nearest anchor} & \multicolumn{3}{c}{Mean CLIP Score} \\
\cmidrule(lr){1-3}\cmidrule(lr){4-6}\cmidrule(lr){7-9}\cmidrule(lr){10-12}
Prompt & Anchor & Cos. & \SysName{} & G. Prior & Interp.
& Prompt & Anchor & Cos. & \SysName{} & G. Prior & Interp. \\
\midrule
Tetromino         & square      & 0.8021 & \textbf{0.2740} & 0.1997 & 0.2242 & Tie              & toe         & 0.8861 & \textbf{0.2339} & 0.1936 & 0.1875 \\
Pikachu           & banana      & 0.8369 & \textbf{0.2354} & 0.1800 & 0.1840 & Box              & book        & 0.8943 & \textbf{0.2770} & 0.2176 & 0.2139 \\
Dancing-human     & hand        & 0.8451 & \textbf{0.2680} & 0.1870 & 0.1959 & a stretching cat & cat         & 0.8971 & \textbf{0.2529} & 0.2248 & 0.2274 \\
Slime             & brain       & 0.8534 & \textbf{0.2310} & 0.1993 & 0.1953 & Ghost            & saw         & 0.8993 & \textbf{0.2449} & 0.1909 & 0.1892 \\
Goblin            & brain       & 0.8537 & \textbf{0.1961} & 0.1770 & 0.1778 & a jumping cat    & cat         & 0.9035 & \textbf{0.2632} & 0.2234 & 0.2423 \\
Kirby             & pig         & 0.8570 & \textbf{0.2658} & 0.2022 & 0.1915 & Earth            & saw         & 0.9063 & \textbf{0.2395} & 0.2065 & 0.2057 \\
Mario             & angel       & 0.8620 & \textbf{0.2001} & 0.1672 & 0.1753 & Demon            & dragon      & 0.9072 & \textbf{0.2107} & 0.1839 & 0.1835 \\
Sonic             & saw         & 0.8688 & \textbf{0.2420} & 0.1892 & 0.1941 & Heart            & brain       & 0.9078 & \textbf{0.2526} & 0.1937 & 0.2034 \\
Mickey            & mouse       & 0.8842 & \textbf{0.2738} & 0.1954 & 0.1999 & Spiral           & circle      & 0.9116 & \textbf{0.2764} & 0.2020 & 0.1980 \\
Kinabalu          & mountain    & 0.7458 & \textbf{0.2498} & 0.1888 & 0.1932 & Star of David    & star        & 0.9121 & \textbf{0.2804} & 0.2369 & 0.2238 \\
Smiley sunflower & smiley face & 0.8537 & \textbf{0.2691} & 0.1785 & 0.1815 & a running cat    & cat         & 0.9121 & \textbf{0.2770} & 0.2269 & 0.2419 \\
Zombie            & angel       & 0.8667 & \textbf{0.2030} & 0.1901 & 0.1838 & Sun \& cloud     & cloud       & 0.9133 & \textbf{0.2344} & 0.2038 & 0.1987 \\
White             & star        & 0.8669 & \textbf{0.2247} & 0.2064 & 0.2109 & Ship             & cruise ship & 0.9155 & \textbf{0.2755} & 0.2218 & 0.2278 \\
Pine              & tree        & 0.8690 & \textbf{0.2584} & 0.2192 & 0.2197 & Angry face       & face        & 0.9156 & \textbf{0.2599} & 0.2072 & 0.2080 \\
Letter A          & key         & 0.8698 & \textbf{0.2488} & 0.2245 & 0.2222 & Fire             & saw         & 0.9175 & \textbf{0.2617} & 0.1945 & 0.1986 \\
Potion            & brain       & 0.8747 & \textbf{0.2105} & 0.1943 & 0.1892 & Water            & river       & 0.9186 & \textbf{0.2418} & 0.2236 & 0.2243 \\
Torch             & lighter     & 0.8799 & \textbf{0.2530} & 0.1993 & 0.2054 & Moon \& star     & moon        & 0.9343 & \textbf{0.2394} & 0.1950 & 0.1897 \\
Black             & saw         & 0.8814 & \textbf{0.2143} & 0.1854 & 0.1900 & Fear face        & face        & 0.9352 & \textbf{0.2230} & 0.2095 & 0.2132 \\
Pizza clock       & pizza       & 0.8818 & \textbf{0.2991} & 0.2313 & 0.2133 & Happy face       & smiley face & 0.9365 & \textbf{0.2642} & 0.2450 & 0.2500 \\
Twins             & saw         & 0.8827 & \textbf{0.2056} & 0.1522 & 0.1422 & Sad face         & face        & 0.9366 & \textbf{0.2580} & 0.2094 & 0.2017 \\
Rocket            & star        & 0.8834 & \textbf{0.2393} & 0.1818 & 0.1782 & Two eyes         & eye         & 0.9372 & \textbf{0.2395} & 0.2087 & 0.2203 \\
Egg               & brain       & 0.8852 & \textbf{0.2798} & 0.2264 & 0.2194 & Lightning cloud  & lightning   & 0.9487 & \textbf{0.2647} & 0.2117 & 0.2152 \\
\midrule
\rowcolor{black!4}
\multicolumn{12}{c}{\textbf{Overall average}\quad \SysName{}: \textbf{0.2480}\qquad Gaussian Prior: 0.2024\qquad Interp CLIP: 0.2034} \\
\bottomrule
\end{tabular*}
}
\end{table*}

\subsection{Comparison with T2I+CLIPasso}

We compare \SysName{} with a two-stage T2I+CLIPasso pipeline on six single-concept prompts from our unseen-label evaluation: \textit{Pikachu}, \textit{Tetromino}, \textit{Goblin}, \textit{Rocket}, \textit{Potion}, and \textit{Ghost}. For each prompt, SDXL Base 1.0~\cite{podell2023sdxl} generates one $1024\!\times\!1024$ target using 50 denoising steps, classifier-free guidance $5.0$, and fixed seed 2026, without a refiner or reranking. We then run the official CLIPasso implementation~\cite{vinker2022clipasso} with 16 strokes and 2,001 optimization iterations for seeds 0, 1000, and 2000, retaining the result with the lowest CLIPasso evaluation loss.

\begin{figure*}[t]
    \centering
    \includegraphics[width=\textwidth]{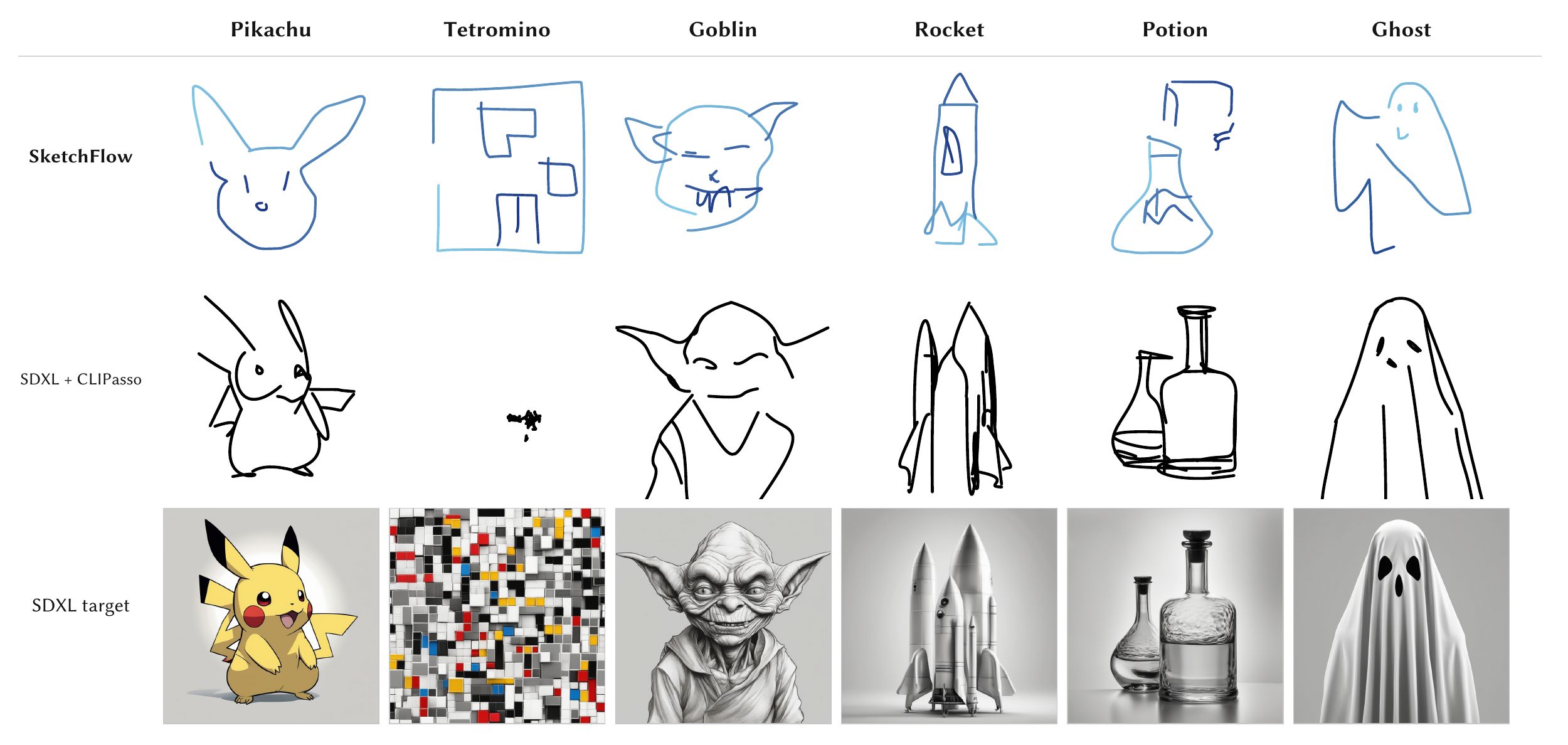}
    \caption{\textbf{Comparison with the two-stage SDXL+CLIPasso baseline.} The \SysName{} row reuses samples already presented in the paper; light-to-dark blue indicates drawing order. The bottom row shows the intermediate SDXL targets supplied to CLIPasso. \SysName{} prioritizes abstract, concept-defining strokes over perspective and volumetric realism, whereas T2I+CLIPasso often preserves more plausible 3D structure from its raster target but may lose simple visual cues essential to conveying the concept.}
    \Description{Qualitative comparison on Pikachu, Tetromino, Goblin, Rocket, Potion, and Ghost. Each column shows a SketchFlow result, the corresponding CLIPasso result, and the intermediate SDXL target image.}
    \label{fig:t2i_clipasso}
\end{figure*}

\subsection{Additional Analysis on NeuralSVG}

NeuralSVG~\cite{polaczek2025neuralsvg} supports controlling its stroke count during optimization. The main paper reports the $32$-stroke setting; here we evaluate $4$, $8$, $16$, and $32$ strokes. Table~\ref{tab:more-quickdraw-results} shows the resulting trade-off: fewer strokes reduce RDP but degrade CLIP alignment, while additional strokes improve CLIP gradually and sharply increase RDP complexity. At 32 strokes, NeuralSVG approaches our CLIP score but retains a much higher RDP value. Our method therefore achieves a more favorable balance between semantic fidelity and geometric efficiency in these comparisons.

\section{User Study Details and Extended Analysis}

As outlined in the main text, we conducted a user study to evaluate the perceptual quality and human-likeness of \SysName. In this supplementary section, we detail the specific questionnaire provided to participants and offer an extended qualitative analysis of the comparative results.

\subsection{Questionnaire Design}
Given the dataset's distinct characteristics and the generation setting beyond a fixed training vocabulary, we refined the evaluation questionnaire to focus heavily on abstraction and structural rationality. Participants were asked to evaluate the generated samples based on the following five specific questions:

{\setlength{\emergencystretch}{1em}
\begin{enumerate}[leftmargin=*, nosep]
    \item \textbf{Abstraction Fidelity:} Which sketch better aligns with the characteristics of an abstract sketch (rather than a traced image)?
    \item \textbf{Semantic Alignment:} Which sketch best resembles the target text prompt?
    \item \textbf{Human-likeness:} Which sketch appears more naturally\linebreak[4] drawn by a human hand?
    \item \textbf{Stroke Rationality:} Which sketch features a more logical and coherent stroke arrangement?
    \item \textbf{Aesthetic Quality:} Which sketch is globally more visually appealing?
\end{enumerate}
}

\subsection{Drawing Order Visualization}
Our framework intrinsically synthesizes sequential vector trajectories. To effectively illustrate the temporal drawing order of the generated sketches throughout this paper, we adopt a color-coding visualization strategy. As depicted in Figure~\ref{fig:exp_draw_sequence}, the temporal progression of individual strokes is represented by varying color intensities, transitioning smoothly from light blue for the initial strokes to dark blue for the final ones. This step-by-step decomposition clearly demonstrates that \SysName{} captures not only the global geometric structure of the target concepts but also learns a coherent drawing sequence that aligns well with plausible human drawing patterns.

\subsection{Extended Qualitative Analysis}
While the quantitative results demonstrate a strong overall preference for \SysName, examining the qualitative traits of the baseline methods provides further context for these scores.

Specifically, successful generation results from NeuralSVG tend to exhibit a dense, image-like quality. They are often characterized by numerous short B\'ezier curves that frequently overlap at the edges, resulting in a visually dense appearance that deviates from concise human abstraction. Conversely, while SketchAgent produces semantically clear outputs, its generated trajectories often appear highly rigid, symmetric, or repetitive, which noticeably diminishes their perceptual human-likeness.

The strong user preference for \SysName\ across all measured metrics is consistent with two traits of our trajectory-aware diffusion approach. First, the generated sketches naturally incorporate authentic drawing nuances (\eg, slight stroke variations), leading to superior human-likeness scores. Second, the evaluated results show abstract expressiveness for challenging, amorphous concepts such as ``water'' and ``fire''. In these examples, the model conveys an abstract impression without being tied to the contours of a specific reference image.

\end{document}